\documentclass[10pt]{article} 

\PassOptionsToPackage{inline}{enumitem}
\usepackage[accepted]{rlj} 

\usepackage{amssymb}            
\usepackage{mathtools}          
\usepackage{mathrsfs}           
\usepackage{graphicx}           
\usepackage[space]{grffile}     
\usepackage{url}                
\usepackage{lipsum}             
\usepackage[utf8]{inputenc} 
\usepackage[T1]{fontenc}    
\usepackage{hyperref}       
\usepackage{url}            
\usepackage{booktabs}       
\usepackage{amsfonts}       
\usepackage{amsmath}        
\usepackage{nicefrac}       
\usepackage{microtype}      
\usepackage{xcolor} 
\usepackage{subfigure}
\usepackage{pdfrender}
\usepackage{multirow}
\usepackage{array, ragged2e, graphicx}
\usepackage{rotating}
\usepackage{caption}
\usepackage{todonotes}
\usepackage{float}
\usepackage{pifont}
\newcommand{\cmark}{{\color{green} \ding{51}}}%
\newcommand{\xmark}{{\color{red} \ding{55}}}%

\usepackage{enumitem}

\title{Assistax: A Multi-Agent Hardware-Accelerated Reinforcement Learning Benchmark for Assistive Robotics}

\setrunningtitle{Assistax}

\author{Leonard Hinckeldey \textsuperscript{1, $\dagger$}, Elliot Fosong \textsuperscript{1, $\dagger$}, Elle Miller \textsuperscript{1}, \\ Rimvydas Rubavicius \textsuperscript{1}, Trevor McInroe\textsuperscript{1}, Fan Zhang \textsuperscript{2}, Patricia Wollstadt \textsuperscript{2}, Stefano V. Albrecht \textsuperscript{3}, Subramanian Ramamoorthy \textsuperscript{1}}

\emails{\{l.hinckeldey, e.fosong, elle.miller, rimvydas.rubavicius, t.mcinroe, s.ramamoorthy\}@ed.ac.uk, \ \ {patricia.wollstadt, fan.zhang}@honda-ri.de}

\affiliations{
$^{1}$\textbf{University of Edinburgh, Edinburgh, UK}\\
$^{2}$\textbf{Honda Research Institute EU, Offenbach am Main, Germany}\\
$^{3}$\textbf{DeepFlow, London, UK}\\
\par 
$^\dagger$ indicating equal contribution
}

\summary{
We introduce Assistax, a benchmark for multi-agent reinforcement learning (MARL) and ad-hoc teamwork (AHT) in assistive robotics. Built entirely in JAX and MuJoCo MJX, it provides end-to-end hardware-accelerated RL training. To the best of our knowledge, Assistax is the first AHT benchmark for 3D continuous environments with two distinct embodied agents in an assistive context. We provide MARL baselines, a diverse population of pre-trained humanoid partners, and an AHT evaluation pipeline that reveals a coordination gap when existing RL algorithms face unseen partners with novel preferences.
}

\contribution{
A fully JAX and MuJoCo MJX-based implementation of five assistive robotics tasks within a scalable benchmark environment, together with a set of tuned MARL baselines that co-train a robot and humanoid agent to solve assistive tasks collaboratively. By leveraging hardware acceleration, Assistax achieves up to 412$\times$ faster open-loop simulation and 25$\times$ faster wall-clock training than comparable CPU-based environments, all on a single GPU. 
}{
JAX based RL environments have become increasingly popular, for MARL JaxMARL~\citep{rutherfordJaxMARLMultiAgentRL2024} provide JAX reimplementations of many existing environment. Few benchmarks offer multi-agent 3D continuous-control (JaxMARL provide MaBrax) tasks at speeds competitive with other MARL environments.  Assistax provides well-motivated tasks of real-world importance in simulation environments that run at the speed of grid-world environments, providing an extensible testbed for RL in assistive robotics.
}\contribution{
An AHT baseline and training pipeline, including pre-training of diverse partner policies using our MARL implementations. These pre-trained partner policies are readily accessible on Hugging Face. We find that existing RL algorithms exhibit a coordination gap when confronted with unseen agents with novel preference combinations, though this gap closes significantly when training and test distributions are similar.
}{
Some of the most popular AHT benchmarks are Overcooked~\citep{carrollUtilityLearningHumans2020a} and Hanabi~\citep{bardHanabiChallengeNew2019}. Neither of these provide a testbed which includes the challenges inherent in assistive robotics. Assistax provides a 3D continuous benchmarking suite that explicitly targets AHT within in an assistive context. We also provide thousands of pre-trained partner policies, for use by AHT and MARL researchers.
}

\keywords{Multi-Agent Reinforcement Learning, Ad-Hoc Teamwork, Zero-Shot Coordination, Assistive Robotics} 

\begin{document}

\makeCover  
\maketitle  

\begin{abstract}
  As embodied autonomous systems capable of assisting humans in daily activities continue to be a major goal for the field of robotics, it becomes increasingly important to have efficient and appropriate reinforcement learning (RL) simulation testbeds. Many common RL environments are too simple to provide insight into complex robotics domains, and many robotics simulations have throughput that is too low for RL applications. In particular, very few robotic simulation environments target multi-agent interactions. While most simulators treat the robot as an isolated agent, real-world tasks such as home assistance and caretaking are inherently multi-agent. Assistax addresses these limitations by providing a high-throughput, scalable suite of GPU-accelerated assistive robotics tasks built on JAX and MuJoCo-MJX, while also including an active humanoid agent serving as a simulated human partner that can be trained alongside the robot using multi-agent RL (MARL). Beyond the application of Assistax as a MARL benchmark environment, we formulate the interaction between the human and the robot as an Ad-Hoc Teamwork (AHT) problem, where the robot's policy must generalise to unseen humans with varying disabilities and preferences. To this end, we provide an extensive AHT benchmarking pipeline: we use MARL to pre-train a diverse population of humanoid partners, and evaluate the ability of the robot policies to coordinate with a withheld set of humanoid policies. In contrast to other benchmarks, we also release reactive MARL-pre-trained humanoid policies via Hugging Face, enabling faster iteration in AHT research. By leveraging hardware acceleration, Assistax achieves up to 412$\times$ faster open-loop simulation than comparable CPU-based environments, all on a single GPU. Our AHT pipeline reveals a coordination gap when we evaluate existing RL algorithms on a set of unseen partners with novel preference combinations. This RL-native test suite for embodied multi-agent interaction provides a practical benchmark for advancing RL in assistive care. The code is readily available on \href{https://github.com/assistive-autonomy/assistax}{GitHub}.
\end{abstract}

\section{Introduction}
\label{Introduction}

Assistive robotics~\citep{6476704,Savage2022RobotsRT} aims to develop autonomous systems that aid humans in performing daily activities, ranging from cleaning and cooking to caring for mobility-impaired individuals in tasks such as bathing and brushing their teeth. A key characteristic of these tasks is that the human is not merely a passive recipient but an active collaborator. In such scenarios, the robot must adapt to the human's behaviour as well as their preferences for how the robot should behave. This robot should also be capable of assisting any potential human; a robot capable of on-the-fly adaptation to any individual user would remove the need for explicit per-user training or fine-tuning.

Beyond dyadic interaction, many of the applications we target require the robot to co-inhabit an environment with multiple diverse agents. Consider a care home setting where the robot must attend not to a single user but to all inhabitants, while also adapting to a constantly changing roster of nurses, doctors, and visiting family members. Such scenarios are often formulated as Ad-Hoc Teamwork (AHT)~\citep{stoneAdHocAutonomous2010}, which requires the robot to coordinate with another agent without any prior experience of that agent. Unlike generalisation across static factors such as physical settings or user appearance, AHT specifically targets adaptation to the \emph{policy} of the partner, a reactive, decision-making agent whose behaviour must be inferred on the fly.


\begin{table}[t]
\centering
\caption{Assistax comparison with related benchmarks. Assistax provides hardware-accelerated, 3D, continuous, multi-agent environments alongside an ad-hoc teamwork (AHT) training pipeline.}
\label{tab:other_benchmarks}
\begin{tabular}{@{}lccccc@{}}
\toprule
\textbf{Benchmarks} &
\textbf{\begin{tabular}[c]{@{}c@{}}Continuous\\actions\end{tabular}} &
\textbf{3D} &
\textbf{\begin{tabular}[c]{@{}c@{}}Hardware\\acceleration\end{tabular}} &
\textbf{\begin{tabular}[c]{@{}c@{}}Multi-\\Agent\end{tabular}} &
\textbf{AHT} \\
\midrule
Bi-DexHands~\citep{chen2022towards}                   & \cmark & \cmark & \cmark & \cmark & \xmark \\
RLBench~\citep{james2019rlbench}                      & \cmark & \cmark & \xmark & \xmark & \xmark \\
Robotsuite~\citep{robosuite2020}                      & \cmark & \cmark & \xmark & \xmark & \xmark \\
Human-Robot Gym~\citep{thummHumanRobotGymBenchmarking2024} & \cmark & \cmark & \xmark & \cmark & \xmark \\
JaxRobotarium~\citep{jain2025jaxrobotariumtrainingdeployingmultirobot} & \cmark & \xmark & \cmark & \cmark & \xmark \\
Assistive Gym~\citep{ericksonAssistiveGymPhysics2019} & \cmark & \cmark & \xmark & \cmark & \xmark \\
JaxMARL~\citep{DBLP:conf/nips/RutherfordEG0LI24}      & \cmark & \cmark & \cmark & \cmark & \xmark \\
Multi-Agent Craftax~\citep{omari2025multiagentcraftaxbenchmarkingopenended} & \xmark & \xmark & \cmark & \cmark & \xmark \\
\midrule
Assistax (ours)                                       & \cmark & \cmark & \cmark & \cmark & \cmark \\
\bottomrule
\end{tabular}
\end{table}

There has been considerable interest in creating autonomous systems that can perform various activities of daily living (ADLs). Initial efforts to standardise research led to the development of frameworks like Assistive Gym~\citep{ericksonAssistiveGymPhysics2019}, which provides a physics-based environment for tasks such as dressing and bathing, and general-purpose manipulation platforms like robosuite~\citep{robosuite2020} that offer high-fidelity control for RL agents. Despite these advancements, few simulation environments actively model the multi-agent interaction between robot and user; for instance, many existing suites treat the human as a static or scripted entity~\citep{thummHumanRobotGymBenchmarking2024} rather than a reactive partner. Furthermore, even fewer such environments are optimised for modern RL research.

Recently, there has been increased interest in hardware-accelerated RL environments that allow for much higher throughput, reducing experiment turnaround to a matter of minutes. In particular, many well-known game and grid-world environments~\citep{samvelyan2019starcraftmultiagentchallenge, lu2022discovered} have been updated to take advantage of modern GPUs, in both RL and MARL~\citep{rutherfordJaxMARLMultiAgentRL2024, gymnax2022github, pignatelli2024navixscalingminigridenvironments}. Similarly, robotics simulations are now leveraging hardware acceleration through tools such as IsaacLab~\citep{mittal2025isaaclab} and MuJoCo MJX~\citep{todorov2012mujoco}, giving rise to a number of accelerated robotics benchmarks, notably in manipulation and locomotion as seen in Bi-DexHands~\citep{chen2022towards} and MuJoCo Playground~\citep{mujoco_playground_2025}.


In light of these advancements, we introduce \textbf{Assistax} --- a hardware-accelerated RL benchmark for assistive robotics. Assistax presents a combination of features that other benchmarks omit (see Table~\ref{tab:other_benchmarks}). Inspired by the recent success of JAX~\citep{jax2018github} in accelerating both RL environments and algorithms, we introduce a suite of MuJoCo MJX-based environments and a range of JAX-implemented algorithms, taking full advantage of hardware acceleration across the entire RL training pipeline. This results in simulation speed-ups of up to 412$\times$ over the closely related, Assistive Gym~\citep{ericksonAssistiveGymPhysics2019} environment. Assistax retains the co-trained humanoid agents of Assistive Gym but extends the evaluation paradigm: rather than testing only with co-trained partners, we evaluate whether robot policies generalise to previously unseen humans with varying preferences, framing the problem as AHT. Our contributions are:


\begin{enumerate}[leftmargin=1.5cm, rightmargin=1.5cm]
    \item A fully JAX and MuJoCo MJX-based implementation of five assistive robotics tasks.
    \item A set of tuned MARL baselines that co-train a robot and humanoid agent to solve assistive tasks collaboratively.
    \item AHT baselines and a training pipeline, including an accessible collection of pre-trained partner policies available on Hugging Face.
\end{enumerate}

\section{Background and Related Work}
\label{sec:background_and_related_wrok}

\textbf{Hardware-Accelerated RL Benchmarks.} Many environments leverage the JAX Python library~\citep{jax2018github} for easy parallelization via \texttt{vmap} (vectorization) and \texttt{pmap} (multi-device distribution) to take full advantage of hardware acceleration~\citep{bettiniVMASVectorizedMultiAgent2022a, DBLP:journals/spe/RichmondCHCL23, mittal2023orbit}. Additionally, MuJoCo's MJX enables collocating agents and environments on GPUs/TPUs, removing CPU-GPU memory transfers and enhancing performance through JAX's JIT compilation. Existing JAX-based RL environments include Gymnax~\citep{gymnax2022github}, Pgx~\citep{koyamada2023pgx}, Navix~\citep{pignatelli2024navixscalingminigridenvironments}, and MuJoCo-based benchmarks~\citep{brax2021github, mujoco_playground_2025}. JaxMARL~\citep{rutherfordJaxMARLMultiAgentRL2024} specifically targets MARL, providing JAX reimplementations for a range of environments and achieving significant speed-ups. Beyond the frameworks mentioned, there are further hardware-accelerated simulation frameworks such as Madrona Engine~\citep{10.1145/3592427, kazemkhani2025gpudrive} and PufferLib~\citep{suarez2024pufferlibmakingreinforcementlearning}.


\textbf{Multi-Agent Reinforcement Learning.} MARL considers multiple RL agents interacting and learning in an environment simultaneously~\citep{marl-book}. The Decentralized Partially Observable Markov Decision Process (Dec-POMDP)~\citep{oliehoekDecentralizedPOMDPFramework2016} is the canonical representation of the decision-making problem in cooperative MARL when agents share the same rewards. It is a tuple $\langle \mathcal{N}, \mathcal{S}, \{ \mathcal{A}^i \}_{i \in \mathcal{N}}, T, p_0, \{ \Omega^i \}_{i \in \mathcal{N}}, O, R, \gamma \rangle$ where: $\mathcal{N}$ is the set of agents; $\mathcal{S}$ is the state space; $\mathcal{A}^i$ is the action space for agent $i$; $T(s_{t+1} \mid s_t, a_t)$ is the state transition probability function; $p_0(s_0)$ is the initial state distribution; $\Omega^i$ is the observation space for agent $i$; $O(o_t \mid s_t, a_{t-1})$ is the observation probability function; $R(s_t, a_t)$ is the team reward function; and $\gamma\in[0,1]$ is the discount factor. In a Dec-POMDP, at every time-step $t$ all agents $i\in\mathcal{N}$ take an action $a_t^i$, each of these actions forms the joint action $a_t = (a_t^1, a_t^2,\dots, a_t^{|\mathcal{N}|})$. After the action $a_t$ is taken, the state transitions to the next state $s_{t+1}$ given by the transition probability function $s_{t+1} \sim T(\cdot \mid s_t, a_t)$. Each agent receives an observation $o_{t+1}^i$ given by the observation probability function $o_t \sim O(\cdot \mid s_t, a_{t-1})$ and receives the reward $r_t$ given by the team reward function $R(s_t, a_t)$. Joint actions are sampled from the joint policy $\pi = (\pi^1, ..., \pi^{|\mathcal{N}|})$ i.e., $a_t \sim \pi(\cdot| h_t)$, where $h_t= (h_t^1, ..., h_t^{|\mathcal{N}|})$ is the joint history. The history of each agent $i \in \mathcal{N}$ consists of all observations and actions up until the current timestep $h^i = (o_0^i, a_0^i,o_1^i, a_1^i,\dots o_t^i)$. The objective is to learn policies that maximize the discounted cumulative rewards over a finite/infinite horizon.

\textbf{Ad-hoc Teamwork.} AHT is the problem of controlling a single agent or a subset of agents within a team, to maximize the team's returns across a broad distribution of previously unknown teammate types~\citep{mirskySurveyAdHoc2022}. A teammate type refers to the distinct set of behavioural policies, capabilities, and preferences that define how a teammate operates and collaborates within a shared task~\citep{albrecht2018modelling,stoneAdHocAutonomous2010}. \textbf{Zero-Shot Coordination (ZSC}), a special case of AHT, requires agents to coordinate with unseen partners without adaptation, typically generalizing from training with a limited set of partner policies~\citep{huOtherPlayZeroShotCoordination2021a}. Existing popular benchmarks for AHT (e.g., Hanabi~\citep{bardHanabiChallengeNew2019} and Overcooked~\citep{carrollUtilityLearningHumans2020a}) often lack the complexity needed for assistive robotics, which involves continuous actions and multi-agent interactions in 3D environments. Assistax explicitly addresses these complexities.
\vspace{-0.5em}
\section{Assistax}
\label{sec:assitax_benchmark}


Assistax is an open-source python library providing 5 assistive robotics tasks, 3 baseline MARL implementations, an Ad-Hoc teamwork baselines, and thousands of pre-trained partner policies. All environments and algorithms are JAX-accelerated, providing a scalable end-to-end training pipeline. 

\begin{figure}[h]
    \centering
    \includegraphics[width=0.8\linewidth]{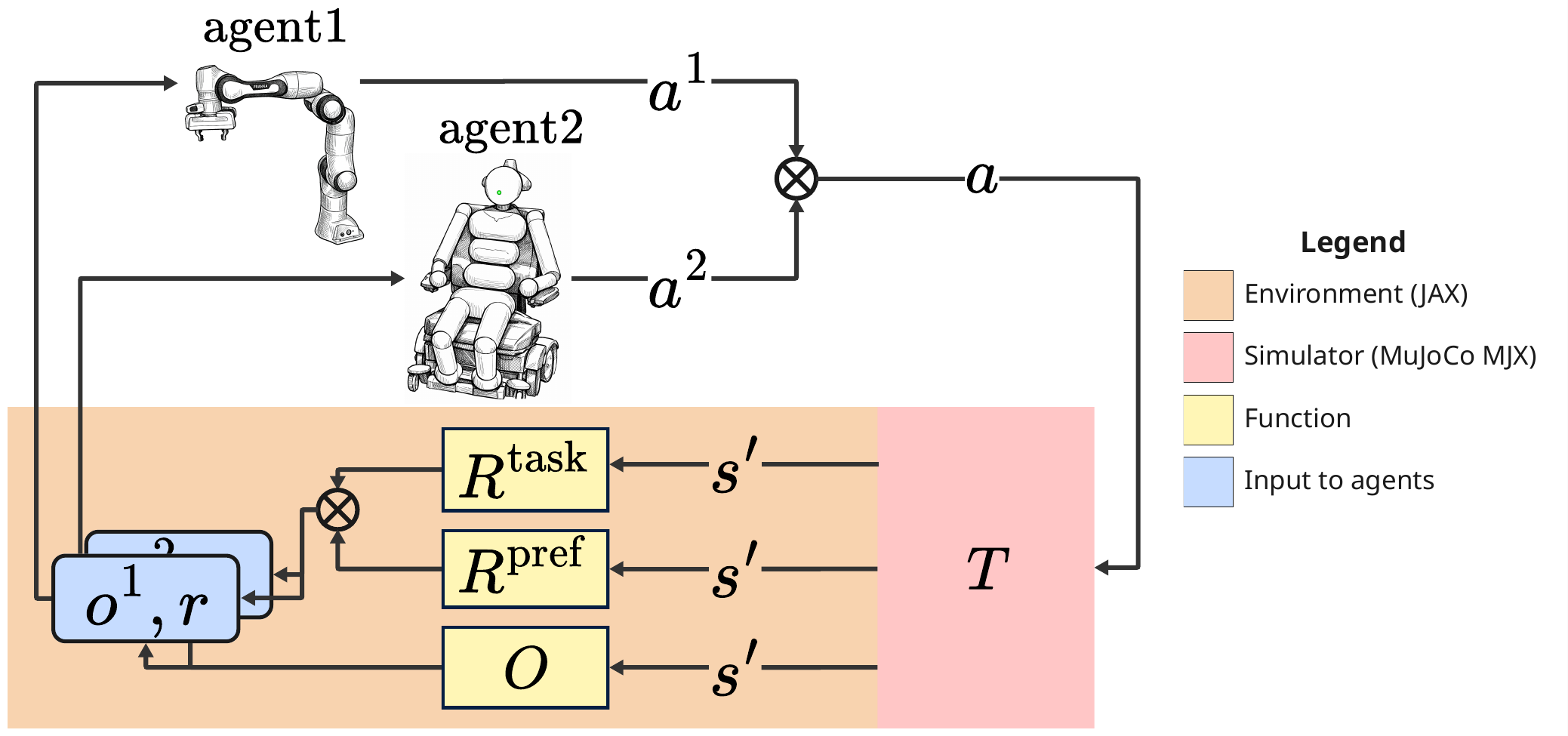}
    \caption{Overview of agent--environment interactions in Assistax. MuJoCo MJX handles state transitions $s' \sim T(\cdot|s, a)$; a JAX-based RL environment computes rewards $r = R(s,a)$ and per-agent observations $o^i \sim O(\cdot|s,a)$. When humanoid preferences are applied, preference rewards augment the task reward: $r = R^\text{task} + R^\text{pref}$. The entire loop can be arbitrarily parallelised with JAX across environments, seeds, hyperparameters, preference settings, and pre-trained humanoids.}
    \label{fig:assistax_loop}
\end{figure}






\subsection{Environments}

\begin{figure*}[h!]
  \centering

    \subfigure[\textit{Scratch}]{\label{fig:scrathitch}\includegraphics[width=0.18\textwidth]{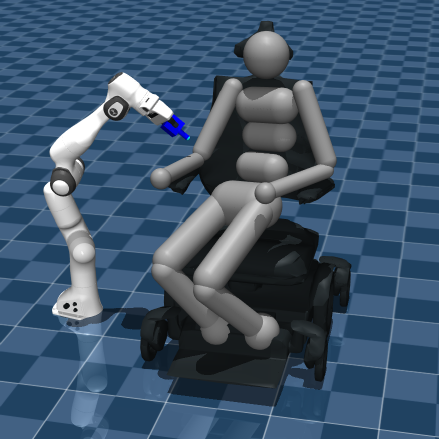}} 
    \subfigure[\textit{Tooth Brushing}\label{fig:teeth}]{\includegraphics[width=0.18\textwidth]{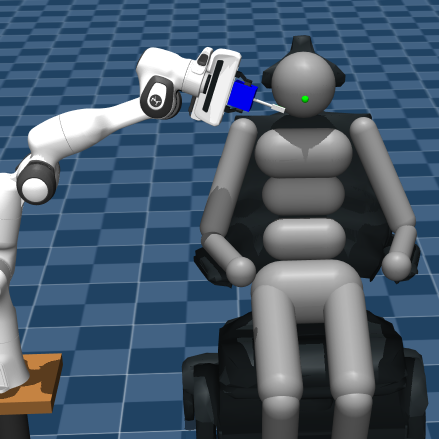}}
    \subfigure[\textit{Feeding}]{\label{fig:feed}\includegraphics[width=0.18\textwidth]{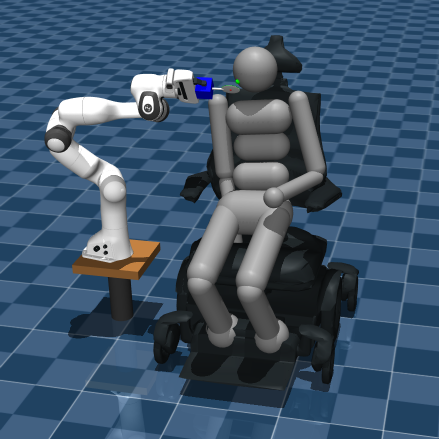}} 
    \subfigure[\textit{Bed Bathing}]
    {\label{fig:bedbathing}\includegraphics[width=0.18\textwidth]{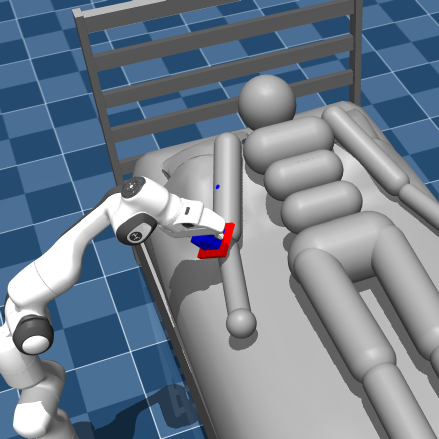}}
    \subfigure[\textit{Arm Assist}]{\label{fig:armassist}\includegraphics[width=0.18\textwidth]{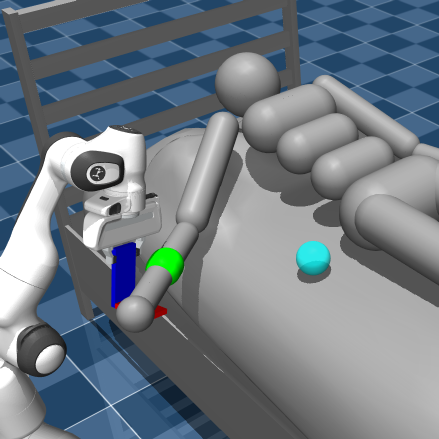}} 
 
    \caption{Assistax environments}
  \label{fig:assistax_tasks}
\end{figure*}

The visualization of the Assistax environment suite is given in Figure~\ref{fig:assistax_tasks}. Inspired by Assistive Gym  \citep{ericksonAssistiveGymPhysics2019}, Assistax implements five environments based on real-world assistive robotic tasks involving collaboration between a human and a robot.

\vspace{0pt}

\begin{itemize}
    \item \textit{Scratch} (Figure~\ref{fig:scrathitch}): A scratching target is randomly sampled on the surface of the humanoid's right arm. The robot must navigate its end-effector to this position and apply a contact force of $3$~N near the target location, while moving its end-effector at $0.1$~m/s to mimic a scratching motion. The scratching reward activates only when the end-effector is close to the target.
    
    \item \textit{Tooth Brushing} (Figure~\ref{fig:teeth}): The robot must navigate the toothbrush to the humanoid's mouth (shown in green in Figure~\ref{fig:teeth}). Rewards are given for proximity to the mouth, alignment of the bristle side of the toothbrush with the mouth, and a brushing component: the Boltzmann product of applying $1$~N of contact force with a tangential speed of $0.1$~m/s. The brushing reward activates only when the toothbrush is within $0.05$~m of the mouth.
    
    \item \textit{Feeding} (Figure~\ref{fig:feed}): The robot must guide a spoon to the humanoid's mouth without spilling. Rewards are given for the distance of the spoon to the mouth, keeping the spoon upright while far from the mouth and tilting it towards the mouth when close, approaching at an appropriate speed, and applying $1$~N of contact force to the mouth.
    
    \item \textit{Bed Bath} (Figure~\ref{fig:bedbathing}): The robot must wipe the humanoid's arm with a sponge. We distribute $52$ target points evenly along the surface of the humanoid's arm. Rewards are given for the distance to the nearest unvisited target and a bonus for each newly visited target. A target counts as visited when the end-effector is within $0.05$~M of it while making contact with the arm. The episode terminates early if all targets are visited.
    
    \item \textit{Arm Assist} (Figure~\ref{fig:armassist}): The robot uses a hook end-effector to lift and reposition the humanoid's weak right arm towards a waist-level target. Rewards are given for the distance between the hook and the wrist target (shown in green in Figure~\ref{fig:armassist}), the distance between the humanoid's lower arm and the waist target (shown as the blue ball in Figure~\ref{fig:armassist}), and the rotational alignment of the hook with the wrist.
\end{itemize}


All tasks use shared rewards between the robot and humanoid agents. Although the rewards target the robot's behaviour, the humanoid can collaborate by making targets more accessible, for example, lifting an arm to expose a scratching target or moving its mouth toward the spoon. Further details are provided in Appendix~\ref{app:benchmark_details}.

\textbf{Agents}. Assistax trains agents that control either a robot or a simplified humanoid model, serving as a simulated human partner. 
For each task, we implement custom end-effectors, which do not have controllable joints. The robot is torque-controlled with 7 joints $\mathcal{A}^\text{robot} \in \mathbb{R}^7 \coloneqq [-1, 1]^7$. The humanoid model is taken from the Brax humanoid tasks~\citep{brax2021github} and is also torque-controlled. For the humanoid we restrict the action space to either the shoulder and elbow joints of the right arm $\mathcal{A}^\text{human} \in\mathbb{R}^3 \coloneqq [-1, 1]^3$ in \textit{Scratch}, \textit{Bed Bath}, and \textit{Arm Assist}; or the yaw and pitch of the humanoid head $\mathcal{A}^\text{human} \in \mathbb{R}^2 \coloneqq [-1, 1]^2$ in \textit{Tooth Brushing} and \textit{Feeding}. Following Assistive Gym, in an attempt to simulate real-world challenges in assistive healthcare tasks, Assistax simulates tremors, joint weakness and limited range of motion as well as a range of preferences regarding how the task is carried out. 

\subsection{Preference Rewards}


To encode a human's preferences for \textit{how} the robot should perform the task, we augment each environment's task reward with a preference reward. Preferences are defined along three axes:
\begin{enumerate*}
\item\textit{force}---the preferred range of contact force the robot applies to the humanoid
\item\textit{speed}---the preferred range of end-effector speed; and 
\item\textit{number of contacts}---a penalty for making new contacts after the initial touch, discouraging poking behaviour in favour of sustained contact (e.g.\ scratching).
\end{enumerate*}
For force and speed, we specify a preferred range and a weighting reflecting how much the humanoid cares about that preference. For contact count, we specify a per-event penalty and an analogous weighting. The three components are normalised into a fixed preference reward budget, calibrated to be between 40\% to 50\% of the maximum attainable per-step reward; each component's share of this budget is proportional to its weighting. When using preference rewards we provide the preferences in the observations for both agents. For further details see Appendix~\ref{app:pref_rew}.

\subsection{Multi-Agent Reinforcement Learning Baselines}
\label{sec:algorithms}

For our MARL baselines we build upon JaxMARL~\citep{rutherfordJaxMARLMultiAgentRL2024} for straightforward single-file implementations. We provide baselines for multi-agent variants of the widely-used proximal policy optimization (PPO)~\citep{schulmanProximalPolicyOptimization2017} and soft-actor critic (SAC)~\citep{haarnojaSoftActorCriticPolicy2018} algorithms, including independent (IPPO) and centralised critic (MAPPO, MASAC) variants. We also include GRU-based~\citep{cho2014propertiesneuralmachinetranslation} recurrent  variations for PPO-based algorithms. Using these baselines we co-train agents to complete the various tasks, i.e. maximizing the expected returns when interacting with the environment as shown in Figure~\ref{fig:assistax_loop}.

\subsection{Ad-Hoc Teamwork Baselines}
\label{sec:aht}

Our AHT pipeline consists of utilising the basic Assistax training loop (as shown in Figure~\ref{fig:assistax_loop}) to pre-train a diverse population of 630 humanoid agents per task. The diversity in the population is induced along three axes:
\begin{enumerate*}
    \item Over-fitting to \textbf{specialized behaviour conventions} which emerge through co-training of agents~\citep{lupuTrajectoryDiversityZeroShot};
    \item varying the \textbf{preferences} of the humanoid when training the population; and
    \item A variety of different \textbf{disability} setting for humanoids in the population. 
\end{enumerate*}


When training the agent population we sample 610 preference combinations uniformly from a constrained set of possible preferences, for more details see the appendix~\ref{app:pref_rew}. We further use 7 distinct disability combinations targeting the mobility of the arm in the relevant tasks. We later discard the robot policies keeping only the pre-trained humanoid policies.

Assistax benchmarks AHT capabilities by considering a simple meta-learning baseline. Using our pre-trained population ($\Pi$) we create disjoint train $\Pi^\mathrm{train}$ and test $\Pi^\mathrm{test}$ sets sampled from $\Pi$. This means that only the robot is learning, while the humanoid follows it's pre-trained policy. Formally, we train the robot for the following objective \citep{rahman2024minimumcoveragesetstraining}:


 \begin{equation*}
    \pi^{i^*}(\Pi^{\mathrm{train}}) = \arg\max_{\pi^i} \mathbb{E}_{\pi^{-i} \sim \mathbb{U}(\Pi^\mathrm{train}), a^i_t \sim \pi_i, a_t^{-i} \sim \pi^{-i}, T, O} \left[\sum_{t=0}^\infty \gamma^tR(a_t, s_t, a_{t+1})\right].
\end{equation*}
 
We then evaluate the measure $M_{\Pi^\mathrm{test}}$  which measures the robustness of $\pi^{i^*}$ when paired with unseen agents uniformly sampled from  $\Pi^\mathrm{test}$, defined as:

\begin{equation*}
    M_{\Pi^\mathrm{test}} = \mathbb{E}_{\pi^{-i} \sim \mathbb{U}(\Pi^\mathrm{test}), a^i_t \sim \pi^{i^*}(\Pi^\mathrm{train}), a_t^{-i} \sim \pi^{-i}, T, O} \left[\sum_{t=0}^\infty \gamma^t R(a_t, s_t, a_{t+1})\right].    
\end{equation*}

Assistax provides the parameters of pre-trained partner policies via Hugging Face, allowing the benchmark users to recreate zero-shot coordination results but also to explore other AHT methods.  

\subsection{Optimized computation}

Assistax prioritises simulation efficiency over high fidelity, a trade-off currently necessary as MuJoCo MJX scales poorly with the number of collisions, resulting in significant slowdowns when simulating deformable bodies such as cloth or detailed meshes. We thus implement solely rigid-body environments using primitive geometries (e.g.\ capsules for the Franka arm, boxes for wheelchairs and beds) and selectively disable collisions between geometries unlikely to interact during an episode.\footnote{These defaults can be adjusted to increase physical fidelity by altering the library's MuJoCo XML files.} While soft-body simulation would be preferable for assistive tasks, Assistax is primarily aimed at researchers in MARL and AHT, who presently often rely on even more simplified settings~\citep{GorsaneMKDSP22}; the growing popularity of JAX-accelerated environments in these communities~\citep{rutherfordJaxMARLMultiAgentRL2024} reflects a desire for high-throughput training that enables fast development and rigorous evaluation. See Figure~\ref{fig:collision_geometries} in Appendix~\ref{app:simulation_fidelity} for further details on collision geometries.

\section{Results}
\label{sec:experiments}

In this section we show the two main results:
\begin{enumerate*}
    \item The MARL performance of robot-humanoid teams co-trained in accordance with Figure~\ref{fig:assistax_loop}.
    \item The AHT setting where the robot learns to complete the task together with a pre-trained humanoid, this policy is trained on a limited train set and evaluated against withheld humanoids, showcasing the zero-shot coordination capabilities of our baselines. 
\end{enumerate*}
For each MARL baseline we conduct extensive hyperparameter tuning, searching over 168 hyperparameter combinations for each baseline--environment pair (without using preference rewards). We use  use the most performant hyperparameter combination found by our tuning, for all our baselines including AHT i.e. using IPPO for the PPO baseline and MASAC values for the SAC baselines. For further details on hyperparameters see Appendix~\ref{sec:hparams}.


\begin{table}[h]
\centering
\caption{Per-episode reward upper bounds for each environment. Task return reflects the task objective alone; the combined column adds the preference reward budget (40--50\% of the overall reward). These bounds are not achievable in practice but serve as a reference for the results in this section.
}
\label{tab:reward_budgets}
\begin{tabular}{lcc}
\toprule
\textbf{Environment} & \textbf{Task Return Max} & \textbf{Combined Max (+ Pref.)} \\
\midrule
Scratching     & 1,000  & 2,000  \\
Bed Bathing    & 1,156  & 2,312  \\
Arm Assist     & 11,280 & 15,792 \\
Feeding        & 4,700  & 6,580  \\
Tooth Brushing  & 3,140  & 4,396  \\
\bottomrule
\end{tabular}
\end{table}

\begin{figure*}[t]
  \centering
  \makebox[\textwidth][c]{%
    \includegraphics[width=0.7\textwidth]{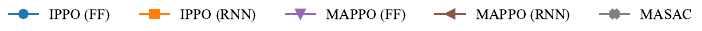}
  }
  \par\vspace{0.5em}
  \subfigure[\textit{Scratching}\label{fig:learning_curve_si}]{
    \includegraphics[width=0.3\textwidth]{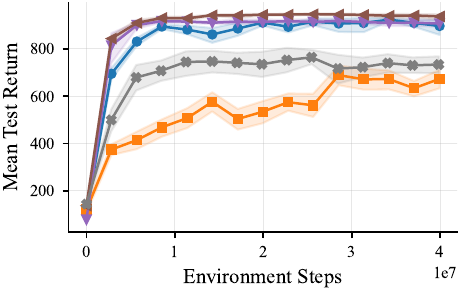}
  }\hfill
  \subfigure[\textit{Bed Bathing}\label{fig:learning_curve_bb}]{
    \includegraphics[width=0.3\textwidth]{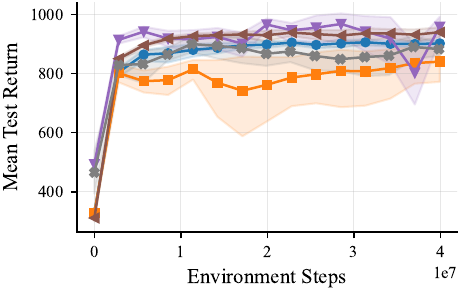}
  }\hfill
  \subfigure[\textit{Arm Assist}\label{fig:learning_curve_aa}]{
    \includegraphics[width=0.3\textwidth]{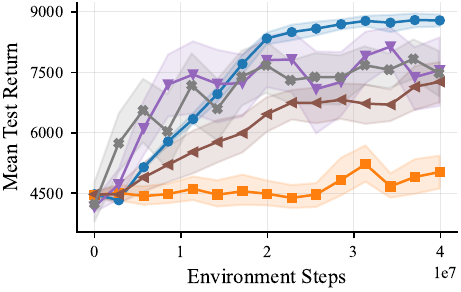}
  }
  \subfigure[\textit{Feeding}\label{fig:learning_curve_feed}]{
    \includegraphics[width=0.3\textwidth]{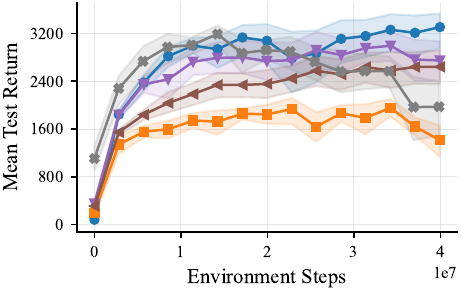}
  }\hfill
  \subfigure[\textit{Tooth Brushing}\label{fig:learning_curve_teeth}]{
    \includegraphics[width=0.3\textwidth]{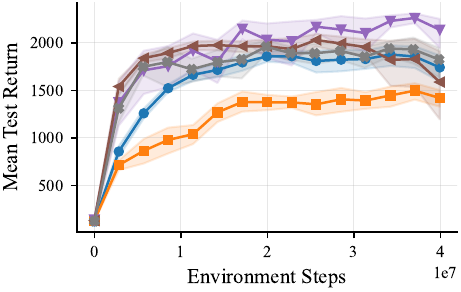}
  }\hfill
  \subfigure[\textit{Aggregate}\label{fig:learning_curve_agg}]{
    \includegraphics[width=0.3\textwidth]{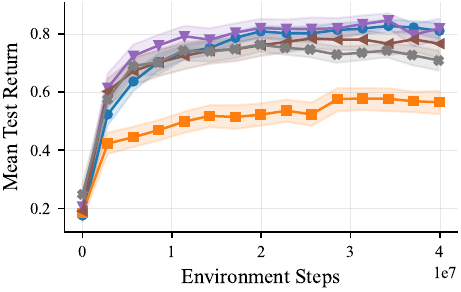}
  }

  \caption{%
    Learning curves for MARL baselines. Each curve shows mean test returns across 16 Seeds and 64 evaluation episodes, and ± 95\% stratified‐bootstrap confidence intervals over those 16 seeds. We also provide Min-Max normalised mean test returns across all environments (see ~\ref{fig:learning_curve_agg}).
  }
  \label{fig:marl_all_tasks_results}
\end{figure*}

\subsection{MARL}

We provide evaluation for four baseline MARL algorithms across all 5 Assistax tasks. We evaluate MASAC, MAPPO and IPPO, each with no parameter sharing \citep{christianos2021scaling}. For MAPPO and IPPO we also provide recurrent baselines. 
 
 

Figure~\ref{fig:marl_all_tasks_results} shows that both IPPO and MAPPO feed-forward implementations consistently rank among the top-performing baselines, with MASAC also performing well in many environments. Despite the partial observability of the environments, recurrent architectures for PPO-based baselines do not outperform their feed-forward counterparts, and actively degrade performance in the case of IPPO. This is in line with results from MABrax~\citep{rutherfordJaxMARLMultiAgentRL2024}, a JAX-accelerated MAMuJoCo benchmark, where feed-forward policies also tend to perform better~\citep{tessera2026probing}.

\begin{figure*}[t!]
  \centering

  \makebox[\textwidth][c]{%
    \includegraphics[width=0.7\textwidth]{figures/shared_legend.pdf}
  }
  \par\vspace{0.5em}

  \makebox[\textwidth][c]{%
    \subfigure[\textit{Without Preferences}\label{fig:learning_curve_nopref}]{
      \includegraphics[width=0.3\textwidth]{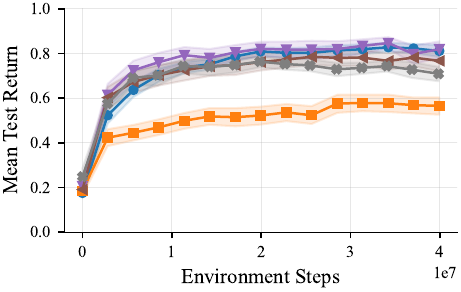}
    }
    \hspace{0.03\textwidth}
    \subfigure[\textit{With Preferences}\label{fig:learning_curve_pref}]{
      \includegraphics[width=0.3\textwidth]{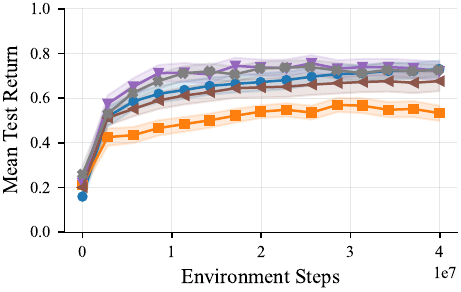}
    }
  }

  \caption{%
    Min-Max normalised test returns and 95\% stratified bootstrapped CI across 16 seeds and all environments. This compares performance with and without additional preference rewards.
  }
  \label{fig:marl_pref_v_nopref}
\end{figure*}

We also compare min-max normalised mean test returns across all tasks comparing the performance with and without preference rewards (Figure~\ref{fig:marl_pref_v_nopref}). We do not note any meaningful difference between in performance of baselines in the two settings. For a more details see Figure~\ref{fig:per_env_prenopref_results} in Appendix~\ref{app:add_results}.


\subsection{Ad Hoc Teamwork}

Assistax can also be used as an AHT benchmark, to test the ability of the robot agent to adapt to previously-unseen humanoid partners.
We provide simple baselines following the metalearning paradigm identified in \cite{mirskySurveyAdHoc2022}---that is, we use a common RL algorithm (PPO or SAC) to train a robot policy with humanoid policies sampled from a training population. This training procedure follows the same principle used to evaluate zero-shot generalisation of RL algorithms in benchmarks like ProcGen~\citep{cobbe2019procgen}, or the domain randomisation techniques employed for sim-to-real transfer in robotics~\citep{8202133}: train and test the algorithm against two distinct sets of partners---a training set and a held-out test set respectively.
To ensure these ad-hoc teamwork benchmarks can be consistently used by other researchers, we make the training and test partner policies available via Hugging Face. 

\begin{figure*}[h!]
  \centering
  \makebox[\textwidth][c]{%
    \includegraphics[width=0.4\textwidth]{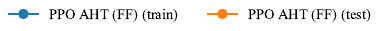}
  }
  \par\vspace{-0.3em}
\subfigure[\textit{Scratching}\label{fig:scratch_aht}]{
    \includegraphics[width=0.3\textwidth]{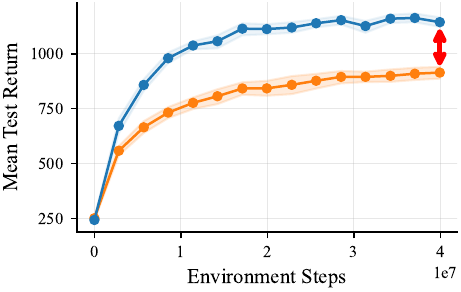}
  }
\subfigure[\textit{Bed Bathing}\label{fig:bedbathing_aht}]{
    \includegraphics[width=0.3\textwidth]{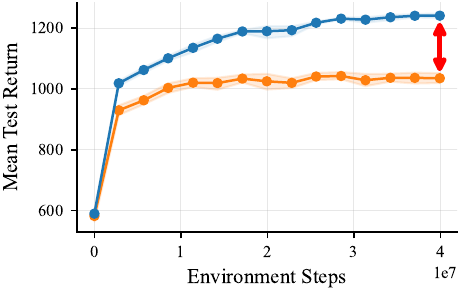}
  }
\subfigure[\textit{Arm Assist}\label{fig:armassist_aht}]{
    \includegraphics[width=0.3\textwidth]{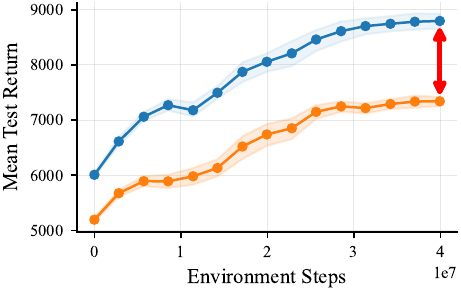}
  }

\subfigure[\textit{Feeding}\label{fig:feeding_aht}]{
    \includegraphics[width=0.3\textwidth]{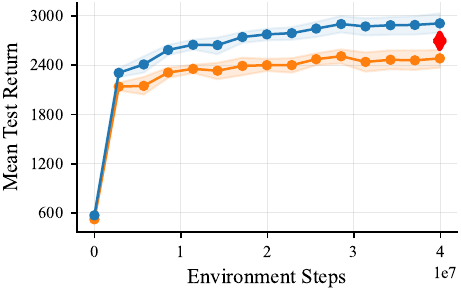}
  }
\subfigure[\textit{Teethbrushing}\label{fig:teeth_aht}]{
    \includegraphics[width=0.3\textwidth]{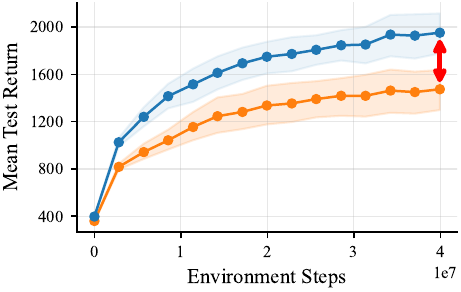}
  }
\subfigure[\textit{Aggregate preference-only returns}\label{fig:agg_aht_pref}]{
    \includegraphics[width=0.3\textwidth]{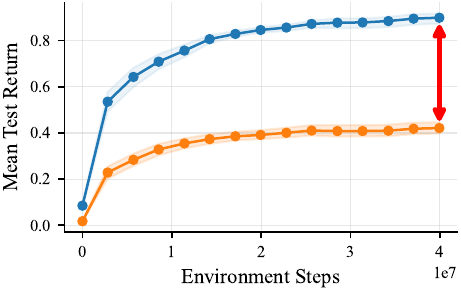}
  }
  \caption{%
    Ad-hoc teamwork performance of PPO across all five tasks. Mean returns and stratified 95\% bootstrapped CI across 16 seeds. Red arrows highlight the coordination gap. 
  }
  \label{fig:aht_results}
\end{figure*}

For the results shown in Figure~\ref{fig:aht_results}, we train against a population of 5 agents and test against 625 unseen agents. We sort agents and select the 5 agents with the highest speed preference weighting. This sampling strategy is deliberately challenging: it minimises the overlap between train and test sets and makes it more likely that the test set includes out-of-distribution preferences. Note that the coordination gap drops significantly when including more agents in the training set or when sampling train and test agents uniformly from the population. In this hard setting, we observe a significant gap in performance that is almost entirely attributable to poor adaptation to new preferences. In Figure~\ref{fig:agg_aht_pref}, we show the min-max normalised mean preference returns (i.e.\ excluding task rewards). A more detailed breakdown can be found in the Appendix~\ref{app:aht_pref_only}. 
\vspace{-1em}
\vspace{-0.2em}
\subsection{Runtime Experiments}
\label{sec:runtime_experiments}
\begin{figure}[H]
    \centering
    \begin{minipage}{0.49\linewidth}
        \centering
        \includegraphics[width=\linewidth]{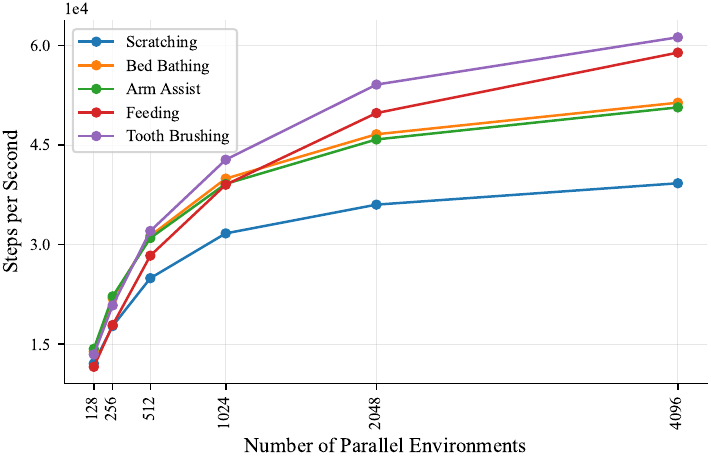}
        \caption{Steps per second and number of vectorized environments using a single A100 (40GB) across Assistax tasks.}
        \label{fig:vec_speed}
    \end{minipage}
    \hfill
    \begin{minipage}{0.49\linewidth}
        \centering
        \captionof{table}{Open-loop simulation speed (steps per second, in hundreds) for Assistax with 1024 vectorised environments on an Nvidia A100 (40GB) GPU, averaged over 10 million timesteps. Relative speed-up is against Assistive Gym on a single AMD EPYC 7763 with no parallel environments.}
        \label{tab:sps_table}
        \begin{tabular}{l
          >{\centering\arraybackslash}p{1.2cm}
          >{\centering\arraybackslash}p{1.2cm}
          >{\centering\arraybackslash}p{1.2cm}}
        \toprule
        \vspace{-1em}
        & \rotatebox{45}{\textbf{Assistax}}
        & \rotatebox{45}{\textbf{Assist. Gym}}
        & \rotatebox{45}{\textbf{Speedup}} \\[1.2em]
        \midrule
        Scratching  & 316.98 & 2.28 & 139$\times$ \\
        Bed Bathing & 399.45 & 0.97 & 412$\times$ \\
        Arm Assist  & 391.54 & 1.40 & 280$\times$ \\
        Feeding     & 390.28 & 1.71 & 228$\times$ \\
        \bottomrule
        \end{tabular}
    \end{minipage}
\end{figure}

A key benefit of implementing the environment in JAX is the significant speed-up in RL training pipelines. To contextualise the results presented here, a typical Assistax IPPO training run of 40 million environment timesteps takes roughly 20 minutes with 1024 parallel environments (see Appendix~\ref{app:runtimes}); the equivalent run in Assistive Gym using RLlib multiprocessing~\citep{liang2018rllibabstractionsdistributedreinforcement} takes approximately 8.3 hours. This yields a roughly 25$\times$ wall-clock reduction, though we note this comparison is indicative rather than controlled---many factors beyond raw simulation speed, such as the RLlib parallelisation strategy used in Assistive Gym, affect end-to-end training time. Figure~\ref{fig:vec_speed} shows how steps per second scale with the number of vectorised environments. Table~\ref{tab:sps_table} reports open-loop simulation speeds compared with Assistive Gym, providing an upper bound on potential speed-ups. Beyond environment throughput, JAX also enables vectorisation across training runs, seeds, and hyperparameter settings, making sweeps and rigorous evaluations far more efficient.

\section{Conclusion}
\label{sec:conclusions}

We introduce Assistax, a hardware-accelerated MARL benchmark for assistive robotics. The presented task suite and experiments demonstrate that continuous, physics-based 3D environments can, from a computational-cost perspective, compete with simpler, game-like settings commonly used in RL, thereby enabling faster research iteration and more thorough evaluations. Beyond MARL, we provide an AHT benchmarking pipeline that offers a suitable testbed for advancing AHT in a range of assistive tasks. We release all environments, algorithms, and pre-trained partner policies to support reproducibility and future work. As assistive robotics grows in societal importance, we hope Assistax can provide a compelling benchmark to the RL community, bringing the computational efficiency of JAX to bear on the multi-agent challenges inherent in human-robot collaboration.




\textbf{Limitations and Future Work.} There is a trade-off between algorithm runtime and environment fidelity, which may come at the cost of accurately replicating real-world scenarios. Furthermore, MARL policies do not capture the full complexity of human behaviour; there thus remains a considerable challenge in bridging the gap between high-throughput simulators and the real-world tasks we are targeting.

\subsubsection*{Acknowledgments}
\label{sec:ack}

Work supported by a UKRI Turing AI World Leading Researcher Fellowship on AI for Person-Centred and Teachable Autonomy (grant EP/Z534833/1). This work was also supported by the Edinburgh International Data Facility (EIDF) and the Data-Driven Innovation Programme at the University of Edinburgh. Access to EIDF was facilitated through the University of Edinburgh’s Generative AI Laboratory GAIL Fellow scheme.

We thank, Christiane Wiebel-Herboth for her contributions to the early stages of this project. We also thank Kale-ab Tessera and Ivalin Chobanov for their insightful discussions.

\appendix

\bibliography{references}
\bibliographystyle{rlj}

\beginSupplementaryMaterials
\appendix

\section{Benchmark Details}
\label{app:benchmark_details} 

\subsection{Observation Space.} 

A summary of the observations for each task is shown in Table~\ref{tab:obs}. In our benchmark, we consider 3 types of observations: proprioception, tactile, and ground-truth information from the simulator. Proprioception is information relating to robot configuration. It is computed from the robot's internal sensors. Assistax considers tactile observation of the net contact forces between the end-effector and the humanoid arm, expressed as a force vector in the MuJoCo contact frame. Observations about the other body forces are not included. Ground truth refers to the privileged information available in simulation but requires estimation in the real world (e.g. end-effector to object distance, humanoid joint angles). Subscripts $R$ and $H$ denote robot and humanoid respectively. The end-effector is chosen to be an imaginary frame at the end of the robotic arm chosen. Note that \textit{Arm Assist} has a larger observation space to take into account the increased task complexity. For this task, we provide the rotation matrix between the end-effector and the target on the humanoid arm (green in Figure~\ref{fig:armassist}) to influence the robot to lift the arm in a particular way. We also provide an additional distance target for the second phase of bringing the arm back to the waist target position (blue in Figure~\ref{fig:armassist}).

\begin{table}[h]
\centering
\caption{Observations space overview. Assistax uses three types of observations: proprioception (prop.), tactile, and ground-truth (gt.) information from the simulator.}
\label{tab:obs}
\begin{tabular}{@{}cclccccccc@{}}
\toprule
\multirow{2}{*}{\textbf{Type}} & \multirow{2}{*}{\textbf{Symbol}} & \multicolumn{1}{c}{\multirow{2}{*}{\textbf{Description}}} & \multirow{2}{*}{\textbf{Dim.}} & \multicolumn{5}{c}{\textbf{Task}} \\
 & & \multicolumn{1}{c}{} & & \textit{\rotatebox{90}{Scratch}} & \textit{\rotatebox{90}{Bed Bath}} & \textit{\rotatebox{90}{Arm Assist}} & \textit{\rotatebox{90}{Feeding}} & \textit{\rotatebox{90}{Teeth Brush}} \\ \midrule
prop. & $\theta_R$ & robot joint angles & 7 & \cmark & \cmark & \cmark & \cmark & \cmark \\
 & $\dot{\theta}_R$ & robot joint velocities & 7 & \cmark & \cmark & \cmark & \cmark & \cmark \\
 & $x_{ee}$ & end-effector position & 3 & \cmark & \cmark & \cmark & \cmark & \cmark \\
 & $q_{ee}$ & end-effector quaternion & 4 & \cmark & \cmark & \cmark & \cmark & \cmark \\ \midrule
tactile & $f_{ee}$ & end-effector force & 3 & \cmark & \cmark & \cmark & \cmark & \cmark \\ \midrule
gt. & $\theta_H$ & humanoid joint angles & 9 & \cmark & \cmark & \cmark & \cmark & \cmark \\
 & $\dot{\theta}_H$ & humanoid joint velocities & 9 & \cmark & \cmark & \cmark & \cmark & \cmark \\
 & $x_{H_{\text{lower}}}$ & humanoid lower arm position & 3 & \cmark & \cmark & \cmark & \xmark & \xmark \\
 & $x_{H_{\text{upper}}}$ & humanoid upper arm position & 3 & \cmark & \cmark & \cmark & \xmark & \xmark \\
 & $x_{H_{\text{mouth}}}$ & humanoid mouth position & 3 & \xmark & \xmark & \xmark & \cmark & \cmark \\
 & $x_{ee\_t}$ & \begin{tabular}[c]{@{}l@{}}end-effector to\\target distance\end{tabular} & 3 & \cmark & \cmark & \cmark & \cmark & \cmark \\
 & $d_{ee\_t}$ & \begin{tabular}[c]{@{}l@{}}end-effector to target\\euclidean distance\end{tabular} & 1 & \cmark & \cmark & \cmark & \cmark & \cmark \\
 & $\mathbf{R}_{ee\_t}$ & \begin{tabular}[c]{@{}l@{}}end-effector to target\\angular distance\end{tabular} & 9 & \xmark & \xmark & \cmark & \cmark & \cmark \\
 & $x_{H\_t'}$ & \begin{tabular}[c]{@{}l@{}}humanoid arm to waist\\target distance\end{tabular} & 3 & \xmark & \xmark & \cmark & \xmark & \xmark \\
 & $d_{H\_t'}$ & \begin{tabular}[c]{@{}l@{}}humanoid arm to waist\\target euclidean distance\end{tabular} & 1 & \xmark & \xmark & \cmark & \xmark & \xmark \\ \bottomrule
\end{tabular}
\end{table}

\begin{table}[H]
\centering
\caption{Reward component overview. Each component is evaluated using the equation and scaled when computing the reward. In equations $\sigma$ is a scaling factor we set to $0.1$, $v_{ee}$ is the end-effector velocity, $[\cdot]$ is the indicator function, $f^*$ and $v^*$ are the target forces and velocities respectively, $v^*_{\text{app}}$ is a distance-dependent target approach speed, and $v_{\tan}$ is the tangential brushing speed.}
\label{tab:reward_components}
\resizebox{\textwidth}{!}{%
\begin{tabular}{@{}rcccccccc@{}}
\toprule
\textbf{Component} & \textbf{Symbol} & \textbf{Equation} & \textbf{Scale} & \multicolumn{5}{c}{\textbf{Task}} \\
\multicolumn{1}{c}{} & & & & \textit{\rotatebox{90}{Scratch}} & \textit{\rotatebox{90}{Bed Bath}} & \textit{\rotatebox{90}{Arm Assist}} & \textit{\rotatebox{90}{Feeding}} & \textit{\rotatebox{90}{Teeth Brush}} \\ 
\midrule
\begin{tabular}[c]{@{}r@{}}Reach\\ target\end{tabular} & $r_t$ & $\exp\!\left(-\frac{d_{ee}^2}{\sigma}\right)$ & 1 & \cmark & \cmark & \cmark & \xmark & \xmark \\
\begin{tabular}[c]{@{}r@{}}Reach\\ target (exp)\end{tabular} & $r_{t_e}$ & $\exp(-3\, d_{ee})$ & 2 & \xmark & \xmark & \xmark & \cmark & \cmark \\
Scratch & $r_s$ & $[d_{ee_t} < 0.1] \cdot \left(\frac{v_{ee}}{v^*}\exp\!\left(-\frac{v_{ee}}{v^*}\right)\right)\cdot\left(\frac{f_{ee}}{f^*}\exp\!\left(-\frac{f_{ee}}{f^*}\right)\right)$ & 4 & \cmark & \xmark & \xmark & \xmark & \xmark \\
Wipe & $r_w$ & $[d_{ee_t} < 0.1] \cdot [f_{ee} > 0]$ & 3 & \xmark & \cmark & \xmark & \xmark & \xmark \\
\begin{tabular}[c]{@{}r@{}}Reach\\ waist\end{tabular} & $r_{t'}$ & $1 - \tanh\!\left(\frac{d_{H_{t'}}}{\sigma}\right)$ & 10 & \xmark & \xmark & \cmark & \xmark & \xmark \\
Rotation & $r_R$ & $\mathrm{norm}(\mathbf{R}_{ee_t})$ & 0.1 & \xmark & \xmark & \cmark & \xmark & \xmark \\
Orientation & $r_o$ & $r_{\text{pour}} + (0.3 + 0.7\cdot p)\, r_{\text{aim}}$ & 1 & \xmark & \xmark & \xmark & \cmark & \xmark \\
\begin{tabular}[c]{@{}r@{}}Approach\\ velocity\end{tabular} & $r_v$ & $\exp\!\left(-\frac{(v_{ee} - v^*_{\text{app}})^2}{0.01}\right)$ & 1 & \xmark & \xmark & \xmark & \cmark & \xmark \\
\begin{tabular}[c]{@{}r@{}}Contact\\ force\end{tabular} & $r_f$ & $\frac{f_{ee}}{f^*}\exp\!\left(-\frac{f_{ee}}{f^*}\right)$ & 1 & \xmark & \xmark & \xmark & \cmark & \xmark \\
\begin{tabular}[c]{@{}r@{}}Alignment\end{tabular} & $r_a$ & $r_{\text{align}} \cdot \exp(-3\, d_{ee})$ & 1 & \xmark & \xmark & \xmark & \xmark & \cmark \\
Brush & $r_b$ & $[d_{ee} < 0.05] \cdot \left(\frac{v_{\tan}}{v^*}\exp\!\left(-\frac{v_{\tan}}{v^*}\right)\right)\cdot\left(\frac{f_{ee}}{f^*}\exp\!\left(-\frac{f_{ee}}{f^*}\right)\right)$ & 1 & \xmark & \xmark & \xmark & \xmark & \cmark \\
\bottomrule
\end{tabular}%
}
\end{table}

\subsection{Simulation Fidelity}
\label{app:simulation_fidelity}

\begin{figure}[H] 
    \centering
    \includegraphics[width=0.3\linewidth]{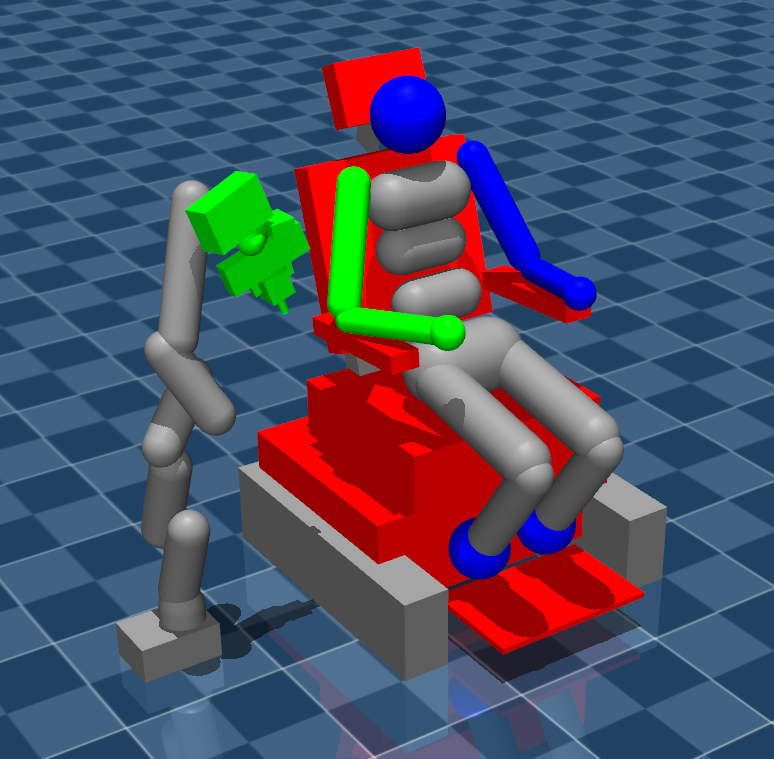}
    \caption{Shows the scratch-itch task with primitive geometries. Green geometries collide with both green and red geometries while blue geometries collide with blue and red geometries. Gray geometries have collision disabled. }
    \label{fig:collision_geometries}
\end{figure}

See Figure~\ref{fig:collision_geometries} for an example of how we set collisions and ue primitive geometries in Assistax and how we use primitive geometries instead of mesh collisions. 

\subsection{Preference Reward System}
\label{app:pref_rew}
The preference rewards augments the task reward with various terms.

\begin{table}[h]
\centering
\caption{Preference reward inputs.}
\label{tab:pref_inputs}
\begin{tabular}{p{0.18\linewidth} p{0.18\linewidth} p{0.56\linewidth}}
\toprule
\textbf{Variable} & \textbf{Units} & \textbf{Meaning} \\
\midrule
\texttt{ee\_speed} & m/s & End-effector speed, $\lVert \mathbf{p}_{t}-\mathbf{p}_{t-1}\rVert / \Delta t$. \\
\texttt{ee\_force} & N & Contact force magnitude on the humanoid, $\lVert \mathbf{f}\rVert$. \\
\texttt{new\_touches} & -- & Penalty for repeated unnecessary contacts, $N_\text{touch}-1$.\\ 
\bottomrule
\end{tabular}
\end{table}

\paragraph{Preference Components.}
The preference reward is a weighted sum of the components in Table~\ref{tab:pref_components}.

\begin{table*}[h]
\centering
\caption{Preference reward components.}
\label{tab:pref_components}
\begin{tabular}{p{0.16\linewidth} p{0.30\linewidth} p{0.17\linewidth} p{0.29\linewidth}}
\toprule
\textbf{Component} & \textbf{Formula} & \textbf{Default} & \textbf{Interpretation / Sweep} \\
\midrule
Speed preference &
$\mathrm{Pref}(v;[v_{\min},v_{\max}])$ &
$[0.06, 0.14]$ m/s &
Rewards motion within the preferred speed range. Sweep: $v_{\min}\sim [0.03,0.08]$, $v_{\max}\sim [v_{\min},0.20]$. \\

Force preference &
$\mathrm{Pref}(f;[f_{\min},f_{\max}])$ &
$[1.5, 3.5]$ N &
Rewards contact force within the preferred range. Sweep: $f_{\min}\sim [1.0,2.0]$, $f_{\max}\sim [f_{\min},5.0]$. \\


Touch penalty &
$\mathbb{1}[f_{t-1}<\tau \;\wedge\; f_t\ge \tau]$ &
weight $=-0.03$ &
Penalizes new touch events. Threshold $\tau=0.1$ N. \\
\bottomrule
\end{tabular}
\end{table*}

\paragraph{Gaussian preference.}
For speed and force, the preference function is
\[
\mathrm{Pref}(x;[x_{\min},x_{\max}]) =
\begin{cases}
1, & x_{\min} \le x \le x_{\max}, \\[4pt]
\exp\!\left(-\dfrac{(x-c)^2}{2w^2}\right), & \text{otherwise},
\end{cases}
\]
where
\[
c=\frac{x_{\min}+x_{\max}}{2},
\qquad
w=\frac{x_{\max}-x_{\min}}{2}.
\]

This yields full reward inside the preferred interval and Gaussian decay outside it.

\paragraph{Budget Normalisation.}
To ensure the total preference reward is calibrated relative to the task reward, we normalise the component weights into a fixed budget. Let $w_s$, $w_f$, and $w_a$ denote the weights for speed, force, and action efficiency respectively, and let $B$ denote the reward budget. We compute the normalisation factor as
\[
\eta = \frac{B}{\,w_s + w_f + w_a\,}.
\]
Each component is then scaled by $\eta$ and its respective weight:
\begin{alignat*}{2}
r_{\text{speed}} &= \eta \, w_s \cdot \mathrm{Pref}(v;\,[v_{\min},v_{\max}]) &\quad &\in \bigl[0,\; \eta \, w_s\bigr], \\
r_{\text{force}} &= \eta \, w_f \cdot \mathrm{Pref}(f;\,[f_{\min},f_{\max}]) &\quad &\in \bigl[0,\; \eta \, w_f\bigr], \\
r_{\text{action}} &= \eta \, w_a \cdot \mathrm{ActionEff}(\mathbf{a}) &\quad &\in \bigl[0,\; \eta \, w_a\bigr], \\
r_{\text{touch}} &= \eta \, w_t \cdot \mathrm{TouchPenalty}(f_t, f_{t-1}) &\quad &\in \bigl[\eta \, w_t,\; 0\bigr],
\end{alignat*}
where $w_t < 0$ is the touch penalty weight. By construction, the maximum positive contribution sums to exactly $B$ regardless of the individual weight values:
\[
\max\bigl(r_{\text{speed}} + r_{\text{force}} + r_{\text{action}}\bigr) = B.
\]
The total preference reward is then
\[
r_{\text{pref}} = w_{\text{overall}} \cdot \bigl(r_{\text{speed}} + r_{\text{force}} + r_{\text{action}} + r_{\text{touch}}\bigr),
\]
where $w_{\text{overall}}$ is a global scaling factor applied to the combined preference signal.

\textbf{During Population Training}

During population training we sample each component uniformly at random, in total we sample 630 preference combinations per task. 

\textbf{Preference Observations}

When preference rewards are enabled we pass the chosen preferences to the agent observations by appending a 7D vector containing the preference ranges and weights. 

\subsection{Baseline Train Times}
\label{app:runtimes}

\begin{table}[ht]
\centering
\caption{Assistax wall-clock IPPO training times. Extrapolated from 5 IPPO updates to full 610 updates utilised when training 40 million timesteps.}
\label{tab:runtime}
\begin{tabular}{lrrrr}
\toprule
& \rotatebox{45}{\textbf{Mean 5 Updates (s)}}
& \rotatebox{45}{\textbf{Std 5 Updates (s)}}
& \rotatebox{45}{\textbf{Est. Full Runtime (s)}}
& \rotatebox{45}{\textbf{Est. Full Runtime (min)}} \\[2.5em]
\midrule
Scratching      & 12.028 & 0.0044 & 1467.42 & 24.46 \\
Bed Bathing     &  9.248 & 0.0025 & 1128.26 & 18.80 \\
Arm Assist      &  9.989 & 0.0034 & 1218.64 & 20.31 \\
Feeding         &  9.602 & 0.0050 & 1171.50 & 19.52 \\
Teethbrushing   &  8.821 & 0.0089 & 1076.18 & 17.94 \\
\bottomrule
\end{tabular}
\end{table}

\section{Additional Results}
\label{app:add_results}

\subsection{Crossplay}

\begin{figure*}[t]
  \centering

  \subfigure[\textit{Teethbrushing}\label{fig:xp_scratchitch}]{
    \includegraphics[width=0.3\textwidth]{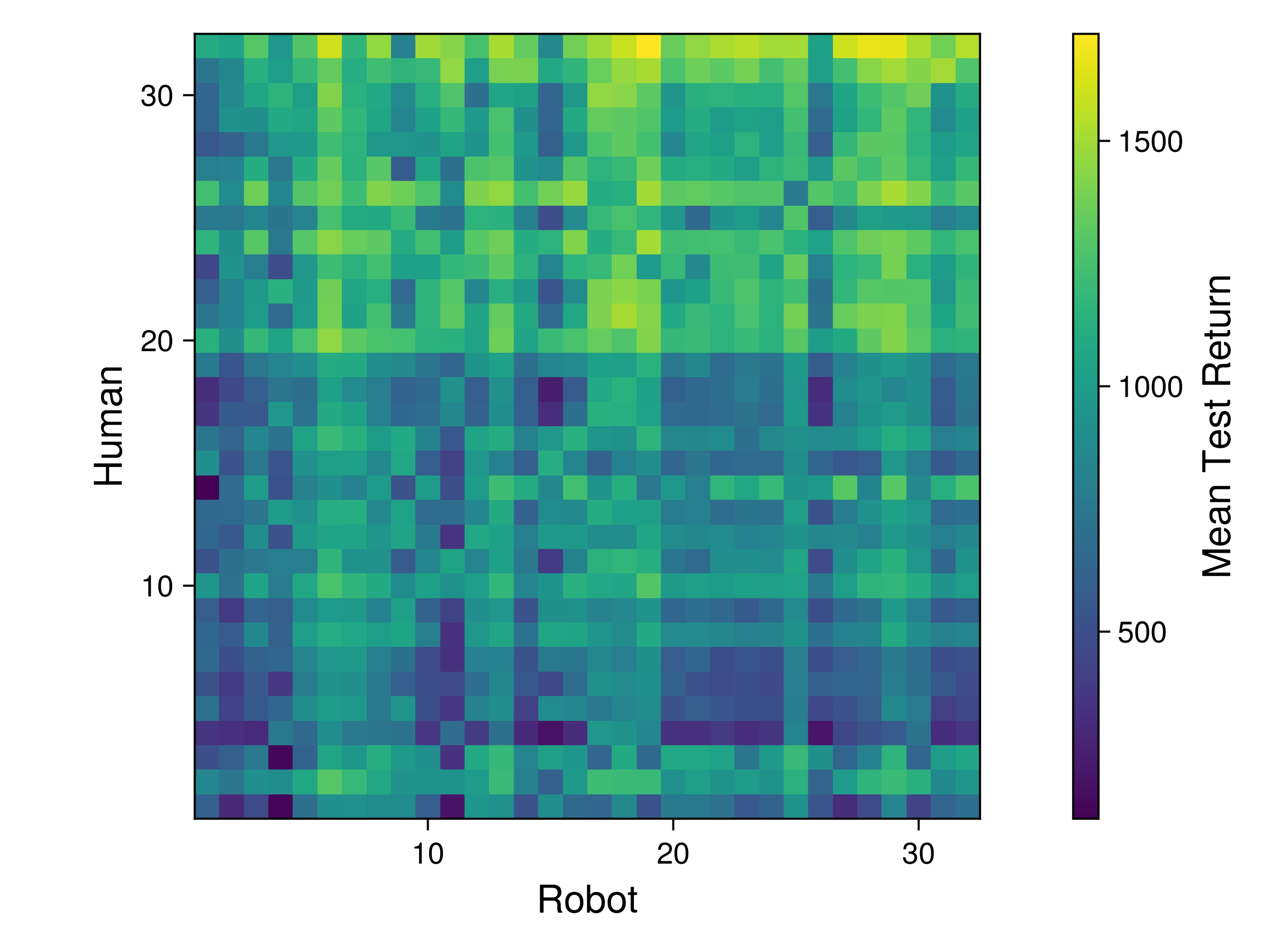}
  }
  \subfigure[\textit{Bed Bathing}\label{fig:crossplay_combined_x}]{
    \includegraphics[width=0.3\textwidth]{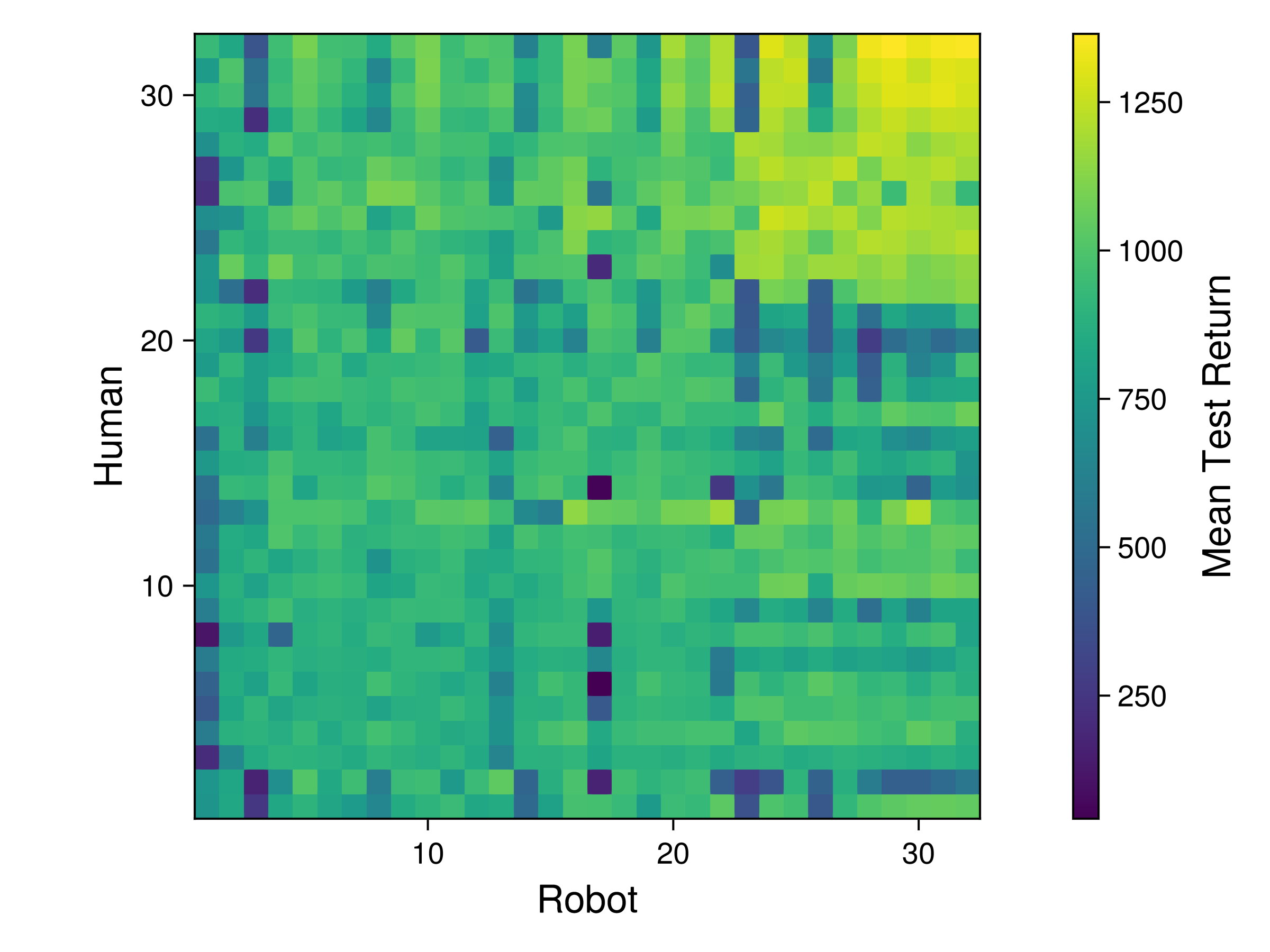}
  }
  \caption{
    Cross-play matrices for a subsample of 32 agents per task. Each cell $(i, j)$ shows the mean test return (averaged over 64 episodes and 16 seeds) when a robot $i$ is paired with humanoid $j$, where the two policies were trained independently in different teams. Co-trained pairs lie on the diagonal. 
  }
  \label{fig:crossplay_matrices}
\end{figure*}

Figure~\ref{fig:crossplay_matrices} show the cross compatibility of random subsamples of our partner population. We note some mutually incompatible strategies emerge notably in Figure~\ref{fig:xp_scratchitch} at index 27, only co-trained agents perform well.  We do not use explicit diversity metrics to encourage diversity in the population beyond preference rewards, future work could consider the use of methods for generating diverse populations of agents such as BRDiv~\citep{rahmanGeneratingTeammatesTraining2023a}.

\subsection{SAC AHT}

\begin{figure*}[h!]
  \centering
  \makebox[\textwidth][c]{%
    \includegraphics[width=0.4\textwidth]{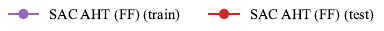}
  }
  \par\vspace{0.5em}
  \subfigure[\textit{Feeding}\label{fig:feeding_aht_sac}]{
    \includegraphics[width=0.3\textwidth]{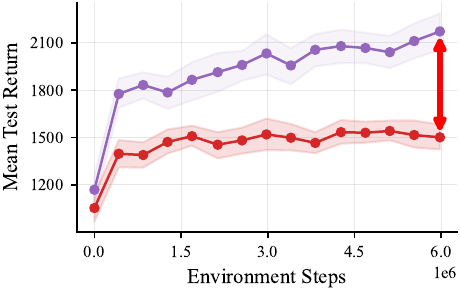}
  }
  \subfigure[\textit{Scratching}\label{fig:scratch_aht_sac}]{
    \includegraphics[width=0.3\textwidth]{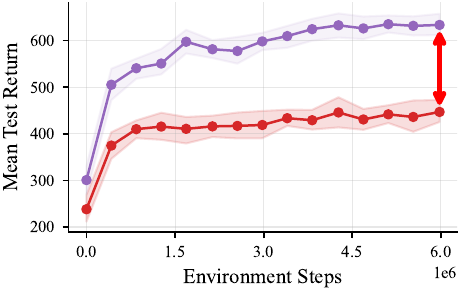}
  }
  \subfigure[\textit{Arm Assist}\label{fig:armassist_aht_sac}]{
    \includegraphics[width=0.3\textwidth]{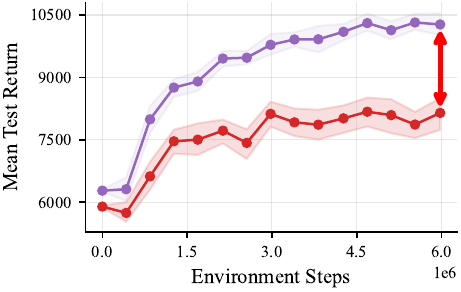}
  }
  \subfigure[\textit{Teethbrushing}\label{fig:teeth_aht_sac}]{
    \includegraphics[width=0.3\textwidth]{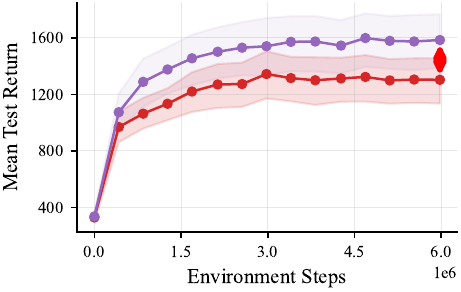}
  }
   \subfigure[\textit{Bed Bathing}\label{fig:bedbathing_aht_sac}]{
    \includegraphics[width=0.3\textwidth]{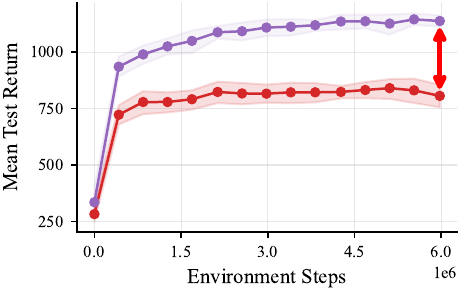}
  }
  \subfigure[\textit{Aggregate preference-only returns}\label{fig:agg_aht_pref_sac}]{
     \includegraphics[width=0.3\textwidth]{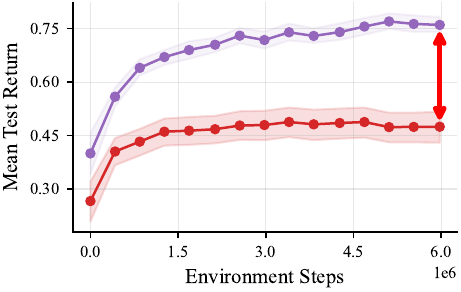}
     }

  \caption{%
    Ad-hoc teamwork performance of SAC across all five tasks. Mean returns and stratified 95\% bootstrapped CI across 16 seeds are shown. Trained for 6 million timestep.
  }
  \label{fig:aht_results_sac}
\end{figure*}

See Figure~\ref{fig:aht_results_sac}, for SAC AHT trained across less timesteps. 

\subsection{AHT Preference Returns Only}
\label{app:aht_pref_only}

\begin{figure*}[h!]
  \centering
  \makebox[\textwidth][c]{%
    \includegraphics[width=0.5\textwidth]{figures/aht_ppo_legend.pdf}
  }
  \par\vspace{0.5em}
  \subfigure[\textit{Feeding}\label{fig:feeding_pref_aht}]{
    \includegraphics[width=0.3\textwidth]{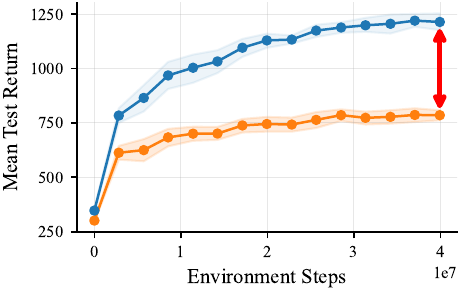}
  }
  \subfigure[\textit{Scratching}\label{fig:scratch_pref_aht}]{
    \includegraphics[width=0.3\textwidth]{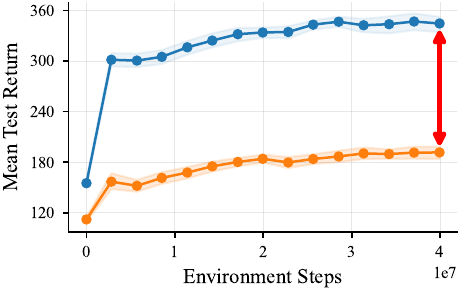}
  }
  \subfigure[\textit{Arm Assist}\label{fig:armassist_pref_aht}]{
    \includegraphics[width=0.3\textwidth]{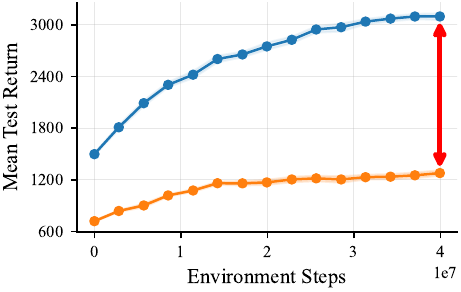}
  }
  \subfigure[\textit{Teethbrushing}\label{fig:teeth_pref_aht}]{
    \includegraphics[width=0.3\textwidth]{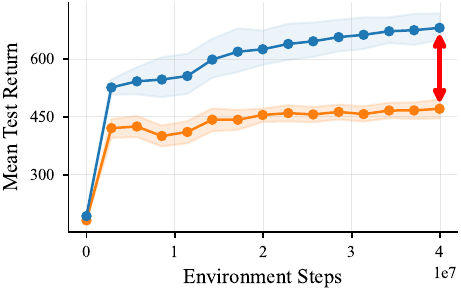}
  }
  \subfigure[\textit{Bed Bathing}\label{fig:bedbathing_pref_aht}]{
    \includegraphics[width=0.3\textwidth]{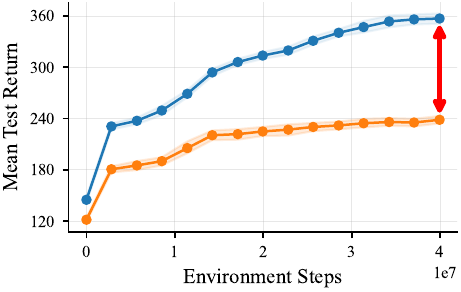}
  }

  \caption{%
Mean preference returns and stratified 95\% bootstrapped CI across 16 seeds are shown. These are returns coming exclusively from the preference reward component. 
  }
  \label{fig:pref_aht}
\end{figure*}

Figure~\ref{fig:pref_aht} shows the coordination gap between only preference returns, excluding task returns. The coordination gap shown in Section~\ref{sec:aht} can attributed almost entirely to poor adaptation to new preferences. 

\subsection{Per Environment Task only vs Task + Preference Rewards}

\begin{figure*}[h!]
  \centering
  \makebox[\textwidth][c]{%
    \includegraphics[width=0.9\textwidth]{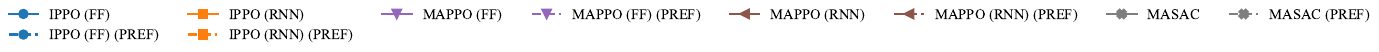}
  }
  \par\vspace{0.5em}
  \subfigure[\textit{Feeding}\label{fig:feeding_pref}]{
    \includegraphics[width=0.3\textwidth]{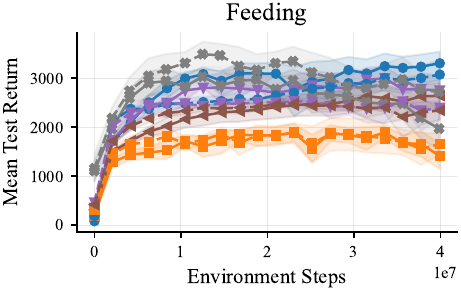}
  }
  \subfigure[\textit{Scratching}\label{fig:scratch_pref}]{
    \includegraphics[width=0.3\textwidth]{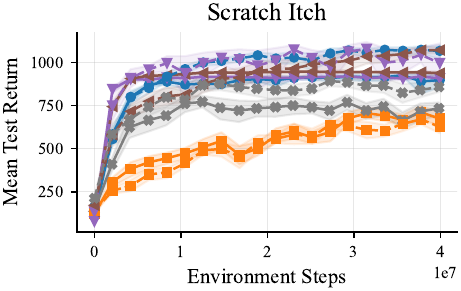}
  }
  \subfigure[\textit{Arm Assist}\label{fig:armassist_pref}]{
    \includegraphics[width=0.3\textwidth]{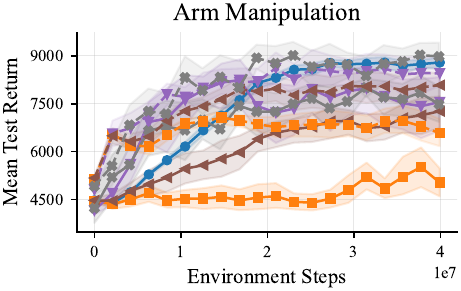}
  }
  \subfigure[\textit{Teethbrushing}\label{fig:teeth_pref}]{
    \includegraphics[width=0.3\textwidth]{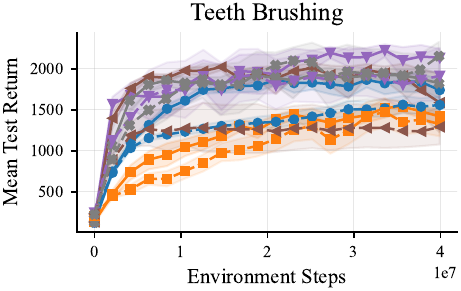}
  }
  \subfigure[\textit{Bed Bathing}\label{fig:bedbathing_pref_comparison}]{
    \includegraphics[width=0.3\textwidth]{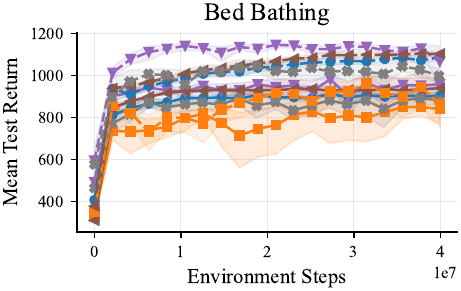}
  }

  \caption{%
Mean returns and stratified 95\% bootstrapped CI across 16 seeds are shown, for environments with and without preference rewards.
  }
  \label{fig:per_env_prenopref_results}
\end{figure*}

See Figure~\ref{fig:per_env_prenopref_results}

\section{Hyperparameters}
\label{sec:hparams}

Below we show the selected hyperparemeters for our baselines. We run one very extensive sweep across our MARL baselines, and reuse these for our AHT Baselines. We provide the same budget of sweeped hyperparemeter i.e. 168 different hyperparameter configurations for all baselines. We evaluete every hyperparameter setting across 12 seeds and 32 independent evaluation episodes. 

We note that since sac has an two additional hyper-parameter which are useful to sweep. This however results in covering less of the search space wihtin each individual hyperparemeter in order to test more different combination of all hyperparemeters. We however feel that by assigning the same compute budget for our sweeps across all algorithms we provide the most fair comparison between them. 
\begin{table*}[h!]
\centering
\caption{Hyperparameters for PPO-based algorithms. PPO\textsubscript{AHT} reuses the IPPO (FF) configuration for each environment. Env.\ abbreviations: SI = Scratching, BB = Bed Bathing, AA = Arm Assist, Feed = Feeding, Teeth = TeethBrushing.}
\label{tab:ppo_hparams}
\resizebox{\textwidth}{!}{%
\begin{tabular}{llccccccc}
\toprule
\textbf{Algorithm} & \textbf{Arch.} & \textbf{Env.} & \textbf{LR} & \textbf{Epochs} & \textbf{Minibatches} & \textbf{Clip $\epsilon$} & \textbf{Ent.\ Coef.} & \textbf{Num Steps} \\
\midrule
\multirow{5}{*}{IPPO}
 & \multirow{5}{*}{FF}
 & SI    & $3.34 \times 10^{-4}$ & 8  & 16 & 0.169 & $1.61 \times 10^{-3}$ & 64 \\
 & & BB    & $8.93 \times 10^{-4}$ & 16 & 8  & 0.042 & $6.38 \times 10^{-5}$ & 64 \\
 & & AA    & $2.45 \times 10^{-4}$ & 16 & 8  & 0.021 & $1.08 \times 10^{-4}$ & 64 \\
 & & Feed  & $3.34 \times 10^{-4}$ & 16 & 16 & 0.169 & $1.61 \times 10^{-3}$ & 64 \\
 & & Teeth & $3.34 \times 10^{-4}$ & 4  & 16 & 0.169 & $1.61 \times 10^{-3}$ & 64 \\
\midrule
\multirow{5}{*}{IPPO}
 & \multirow{5}{*}{RNN}
 & SI    & $4.39 \times 10^{-4}$ & 8  & 4  & 0.046 & $5.45 \times 10^{-4}$ & 64 \\
 & & BB    & $2.89 \times 10^{-3}$ & 16 & 4  & 0.066 & $5.73 \times 10^{-4}$ & 64 \\
 & & AA    & $4.61 \times 10^{-4}$ & 8  & 4  & 0.253 & $1.69 \times 10^{-3}$ & 64 \\
 & & Feed  & $4.61 \times 10^{-4}$ & 8  & 8  & 0.253 & $1.69 \times 10^{-3}$ & 64 \\
 & & Teeth & $1.25 \times 10^{-4}$ & 16 & 16 & 0.130 & $2.39 \times 10^{-3}$ & 64 \\
\midrule
\multirow{5}{*}{MAPPO}
 & \multirow{5}{*}{FF}
 & SI    & $2.56 \times 10^{-3}$ & 8  & 4  & 0.225 & $3.91 \times 10^{-5}$ & 128 \\
 & & BB    & $1.12 \times 10^{-3}$ & 16 & 8  & 0.150 & $2.90 \times 10^{-3}$ & 128 \\
 & & AA    & $1.54 \times 10^{-3}$ & 16 & 4  & 0.271 & $7.95 \times 10^{-4}$ & 128 \\
 & & Feed  & $1.54 \times 10^{-3}$ & 4  & 16 & 0.271 & $7.95 \times 10^{-4}$ & 128 \\
 & & Teeth & $4.61 \times 10^{-4}$ & 8  & 16 & 0.253 & $1.69 \times 10^{-3}$ & 128 \\
\midrule
\multirow{5}{*}{MAPPO}
 & \multirow{5}{*}{RNN}
 & SI    & $4.61 \times 10^{-4}$ & 4  & 8  & 0.253 & $1.69 \times 10^{-3}$ & 128 \\
 & & BB    & $1.04 \times 10^{-4}$ & 16 & 16 & 0.068 & $4.25 \times 10^{-5}$ & 128 \\
 & & AA    & $4.61 \times 10^{-4}$ & 8  & 16 & 0.253 & $1.69 \times 10^{-3}$ & 128 \\
 & & Feed  & $2.56 \times 10^{-3}$ & 8  & 8  & 0.225 & $3.91 \times 10^{-5}$ & 128 \\
 & & Teeth & $2.56 \times 10^{-3}$ & 4  & 8  & 0.225 & $3.91 \times 10^{-5}$ & 128 \\
\midrule
\multirow{5}{*}{PPO\textsubscript{AHT}}
 & \multirow{5}{*}{FF}
 & SI    & $3.34 \times 10^{-4}$ & 8  & 16 & 0.169 & $1.61 \times 10^{-3}$ & 64 \\
 & & BB    & $8.93 \times 10^{-4}$ & 16 & 8  & 0.042 & $6.38 \times 10^{-5}$ & 64 \\
 & & AA    & $2.45 \times 10^{-4}$ & 16 & 8  & 0.021 & $1.08 \times 10^{-4}$ & 64 \\
 & & Feed  & $3.34 \times 10^{-4}$ & 16 & 16 & 0.169 & $1.61 \times 10^{-3}$ & 64 \\
 & & Teeth & $3.34 \times 10^{-4}$ & 4  & 16 & 0.169 & $1.61 \times 10^{-3}$ & 64 \\
\bottomrule
\end{tabular}%
}
\end{table*}

\begin{table*}[h!]
\centering
\caption{Hyperparameters for SAC-based algorithms. SAC\textsubscript{AHT} reuses the MASAC configuration for each environment. All variants use feed-forward architectures. Env.\ abbreviations as in Table~\ref{tab:ppo_hparams}.}
\label{tab:sac_hparams}
\resizebox{\textwidth}{!}{%
\begin{tabular}{lccccccc}
\toprule
\textbf{Algorithm} & \textbf{Env.} & \textbf{Policy LR} & \textbf{Q LR} & \textbf{$\alpha$ LR} & \textbf{$\tau$} & \textbf{SAC Updates} & \textbf{Rollout / Batch} \\
\midrule
\multirow{5}{*}{MASAC}
 & SI    & $5.62 \times 10^{-5}$ & $1.78 \times 10^{-3}$ & $2.52 \times 10^{-4}$ & $2.73 \times 10^{-4}$ & 32  & 8 / 512 \\
 & BB    & $5.62 \times 10^{-5}$ & $1.78 \times 10^{-3}$ & $2.52 \times 10^{-4}$ & $2.73 \times 10^{-4}$ & 64  & 8 / 512 \\
 & AA    & $2.34 \times 10^{-4}$ & $1.37 \times 10^{-5}$ & $1.98 \times 10^{-4}$ & $1.08 \times 10^{-2}$ & 64  & 8 / 512 \\
 & Feed  & $5.62 \times 10^{-5}$ & $1.78 \times 10^{-3}$ & $2.52 \times 10^{-4}$ & $2.73 \times 10^{-4}$ & 128 & 8 / 128 \\
 & Teeth & $5.62 \times 10^{-5}$ & $1.78 \times 10^{-3}$ & $2.52 \times 10^{-4}$ & $2.73 \times 10^{-4}$ & 32  & 8 / 128 \\
\midrule
\multirow{5}{*}{SAC\textsubscript{AHT}}
 & SI    & $5.62 \times 10^{-5}$ & $1.78 \times 10^{-3}$ & $2.52 \times 10^{-4}$ & $2.73 \times 10^{-4}$ & 32  & 8 / 512 \\
 & BB    & $5.62 \times 10^{-5}$ & $1.78 \times 10^{-3}$ & $2.52 \times 10^{-4}$ & $2.73 \times 10^{-4}$ & 64  & 8 / 512 \\
 & AA    & $2.34 \times 10^{-4}$ & $1.37 \times 10^{-5}$ & $1.98 \times 10^{-4}$ & $1.08 \times 10^{-2}$ & 64  & 8 / 512 \\
 & Feed  & $5.62 \times 10^{-5}$ & $1.78 \times 10^{-3}$ & $2.52 \times 10^{-4}$ & $2.73 \times 10^{-4}$ & 128 & 8 / 128 \\
 & Teeth & $5.62 \times 10^{-5}$ & $1.78 \times 10^{-3}$ & $2.52 \times 10^{-4}$ & $2.73 \times 10^{-4}$ & 32  & 8 / 128 \\
\bottomrule
\end{tabular}%
}
\end{table*}

\clearpage
\subsection{Scratching Environment Sweep Plots}

Here we visualize hyperparemeter sweeps for the scratching environment (note that we do sweep across all environments). We provide grids of sweeps where each subplot shows a combination of discrete hyepr parameters which with a random sweep over a continuous hyperparameters. We have 3 grids per algorithm environment pair 1 grid plot for each contionous hyperparameter. 

\begin{figure}[H]
    \centering
    \includegraphics[width=\linewidth]{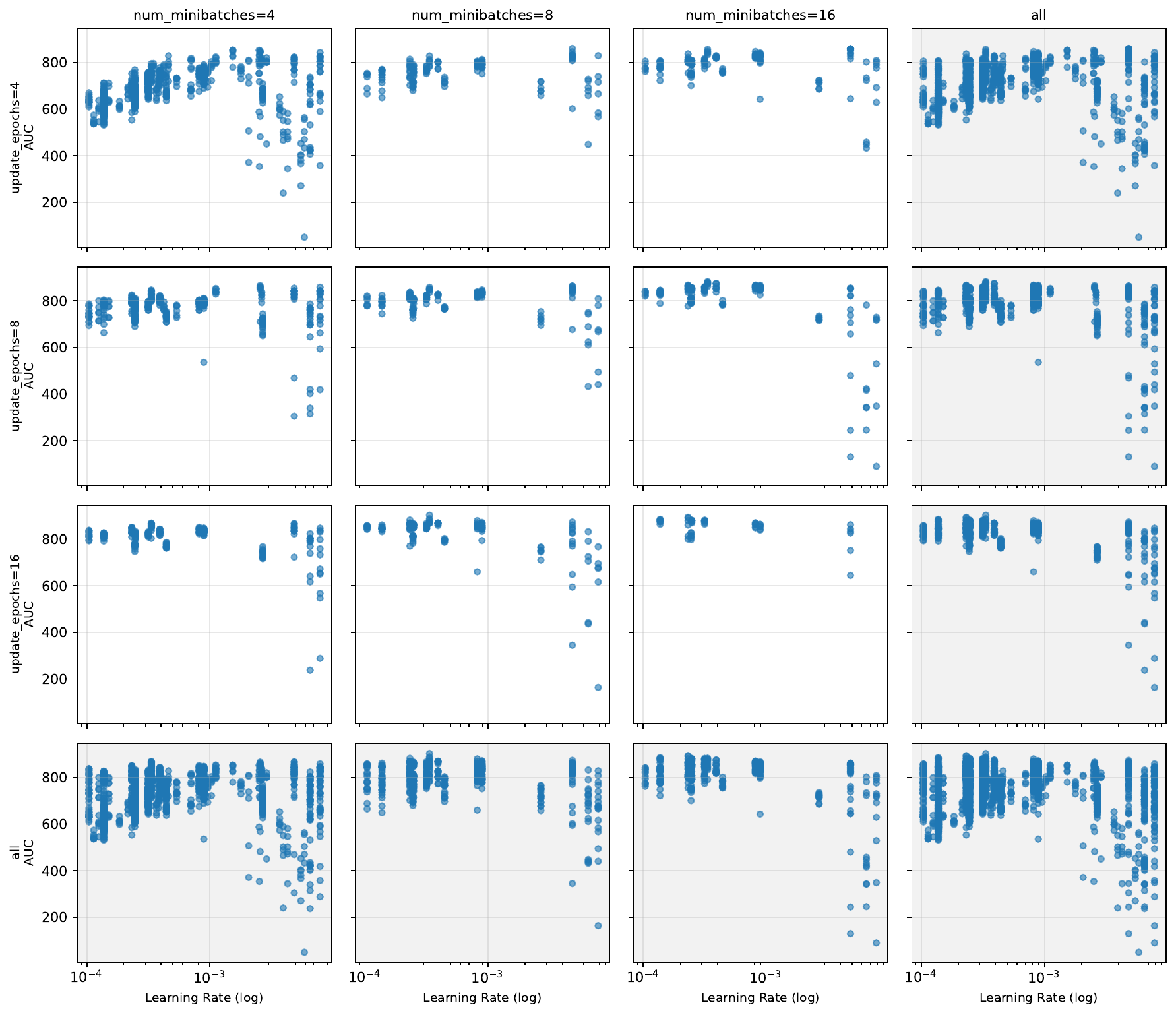}
    \caption{\textbf{IPPO (feed-forward), Scratch Itch --- learning-rate sweep.} Each marker gives one configuration's AUC return as a function of the learning rate (log $x$-axis); higher is better. Panels are faceted by minibatch count (columns) and update epochs (rows), with \texttt{all} pooling over that factor.}
    \label{fig:sweep_ippo_ff_si_lr}
\end{figure}

\begin{figure}[H]
    \centering
    \includegraphics[width=\linewidth]{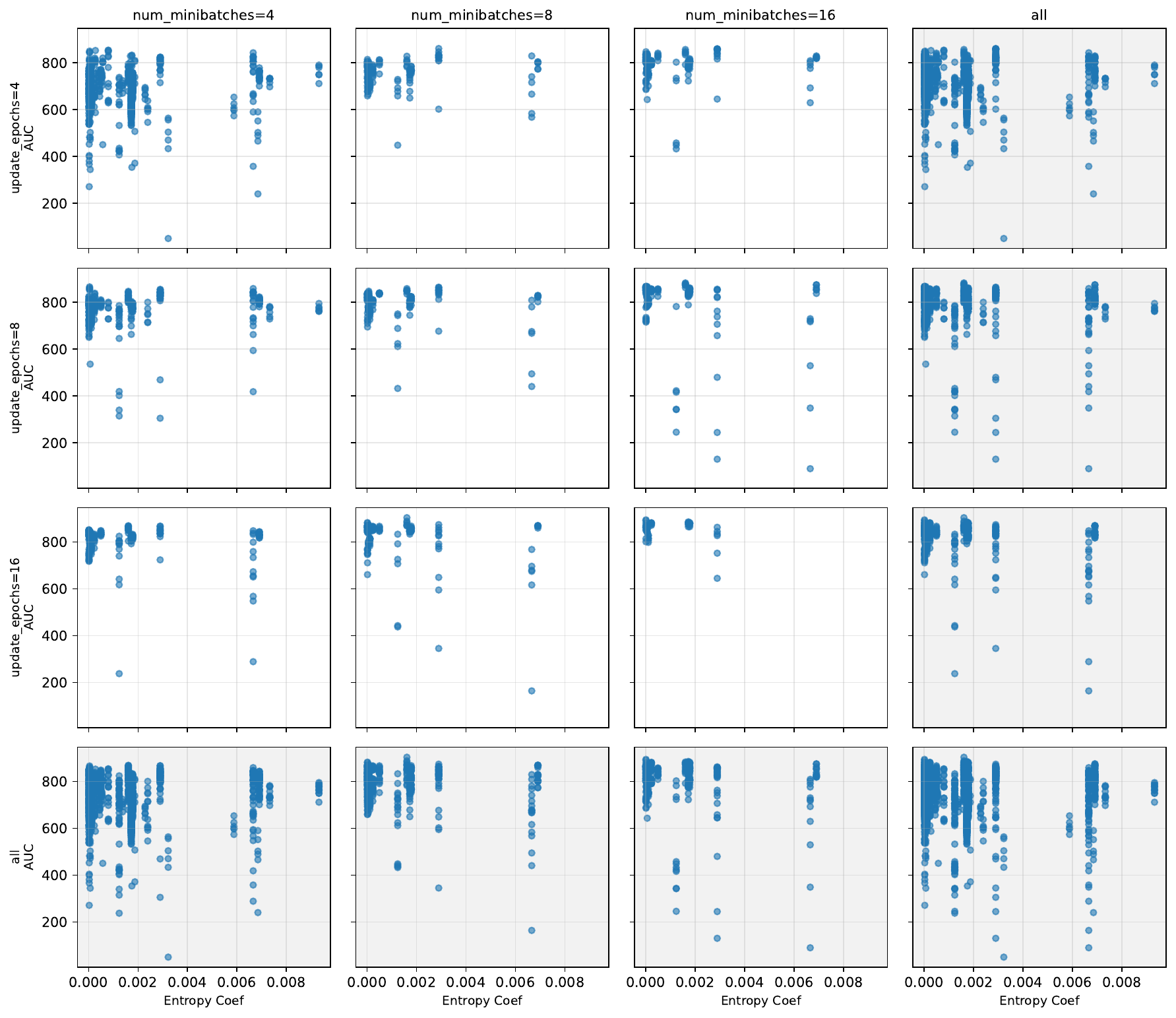}
    \caption{\textbf{IPPO (feed-forward), Scratch Itch --- entropy-coefficient sweep.} AUC return against the entropy-regularisation coefficient, where each marker is a single configuration. Panels are faceted by minibatch count (columns) and update epochs (rows), with \texttt{all} pooling over that factor.}
    \label{fig:sweep_ippo_ff_si_ent}
\end{figure}

\begin{figure}[H]
    \centering
    \includegraphics[width=\linewidth]{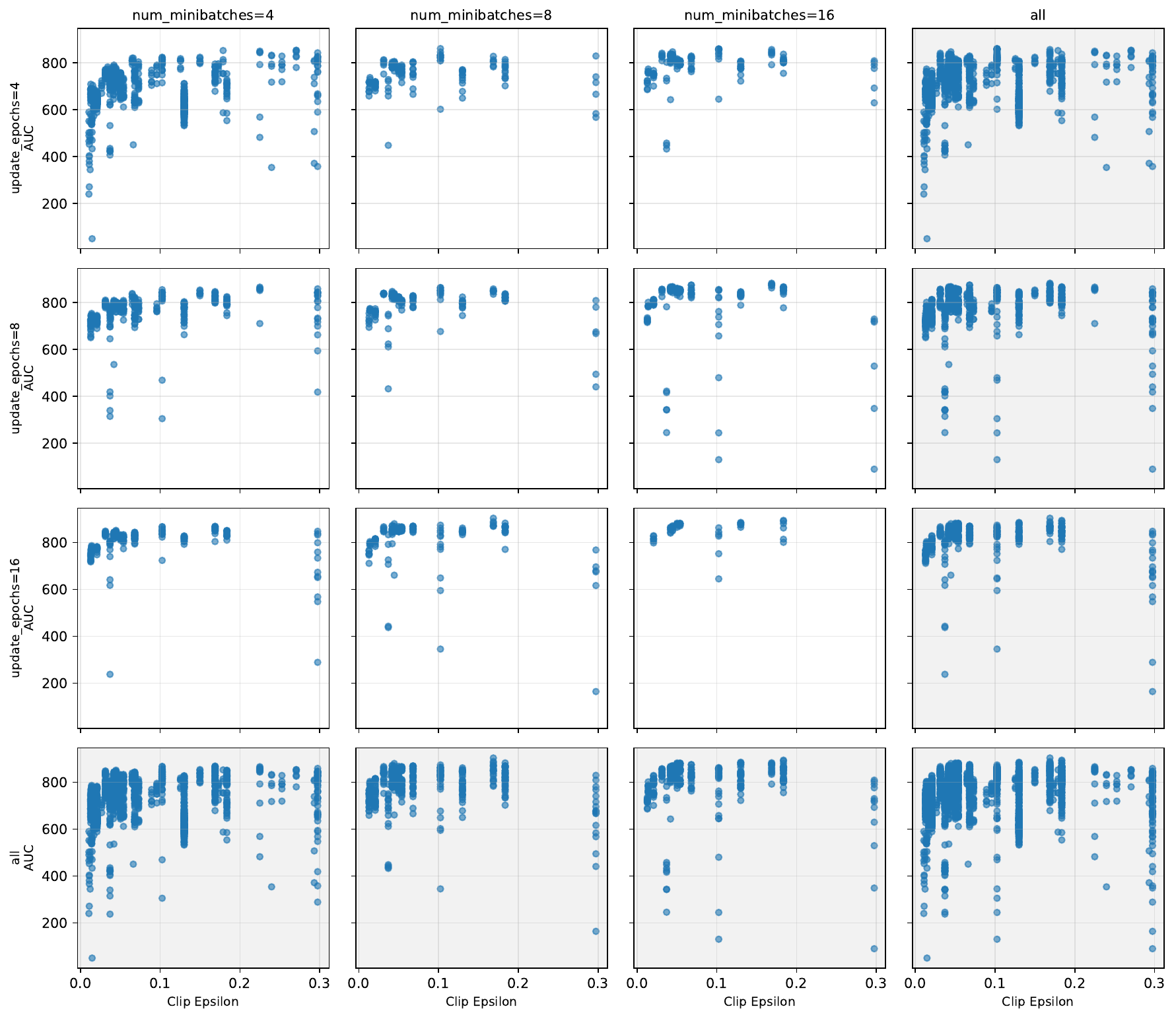}
    \caption{\textbf{IPPO (feed-forward), Scratch Itch --- PPO clip-range sweep.} AUC return against the PPO clipping parameter $\epsilon$, where each marker is a single configuration. Panels are faceted by minibatch count (columns) and update epochs (rows), with \texttt{all} pooling over that factor.}
    \label{fig:sweep_ippo_ff_si_clip}
\end{figure}

\begin{figure}[H]
    \centering
    \includegraphics[width=\linewidth]{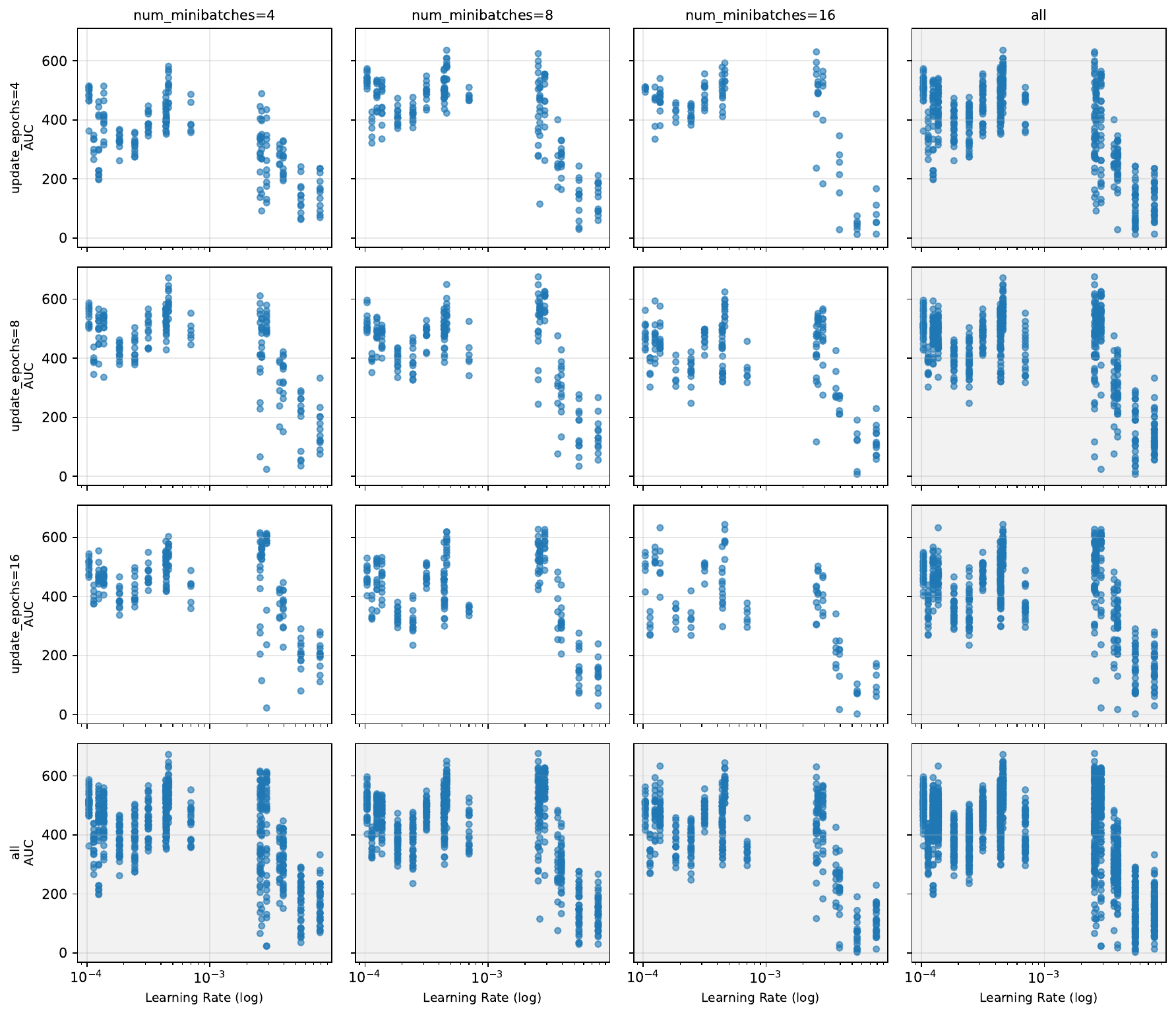}
    \caption{\textbf{IPPO (recurrent), Scratch Itch --- learning-rate sweep.} Each marker gives one configuration's AUC return as a function of the learning rate (log $x$-axis); higher is better. Panels are faceted by minibatch count (columns) and update epochs (rows), with \texttt{all} pooling over that factor.}
    \label{fig:sweep_ippo_rnn_si_lr}
\end{figure}

\begin{figure}[H]
    \centering
    \includegraphics[width=\linewidth]{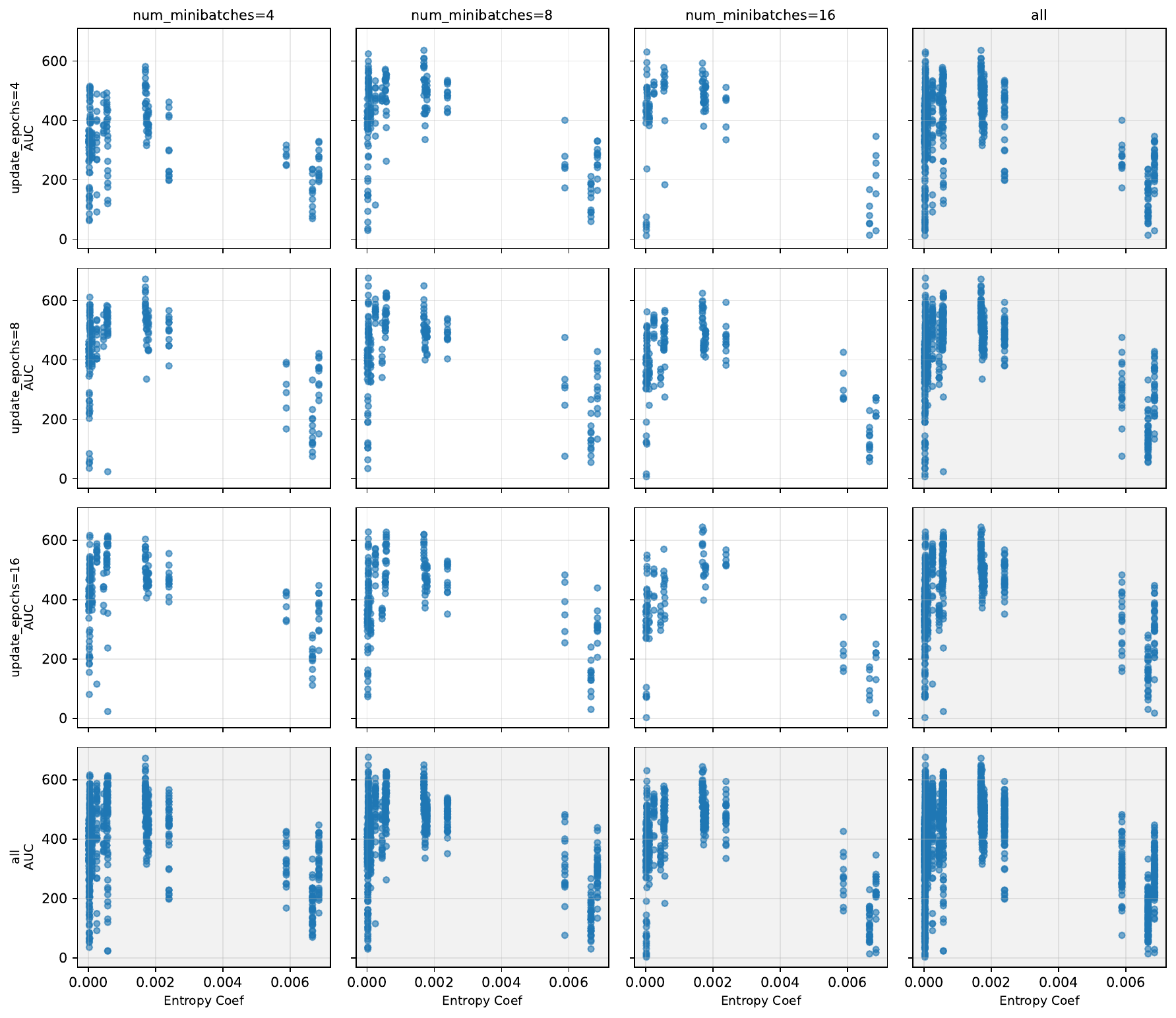}
    \caption{\textbf{IPPO (recurrent), Scratch Itch --- entropy-coefficient sweep.} AUC return against the entropy-regularisation coefficient, where each marker is a single configuration. Panels are faceted by minibatch count (columns) and update epochs (rows), with \texttt{all} pooling over that factor.}
    \label{fig:sweep_ippo_rnn_si_ent}
\end{figure}

\begin{figure}[H]
    \centering
    \includegraphics[width=\linewidth]{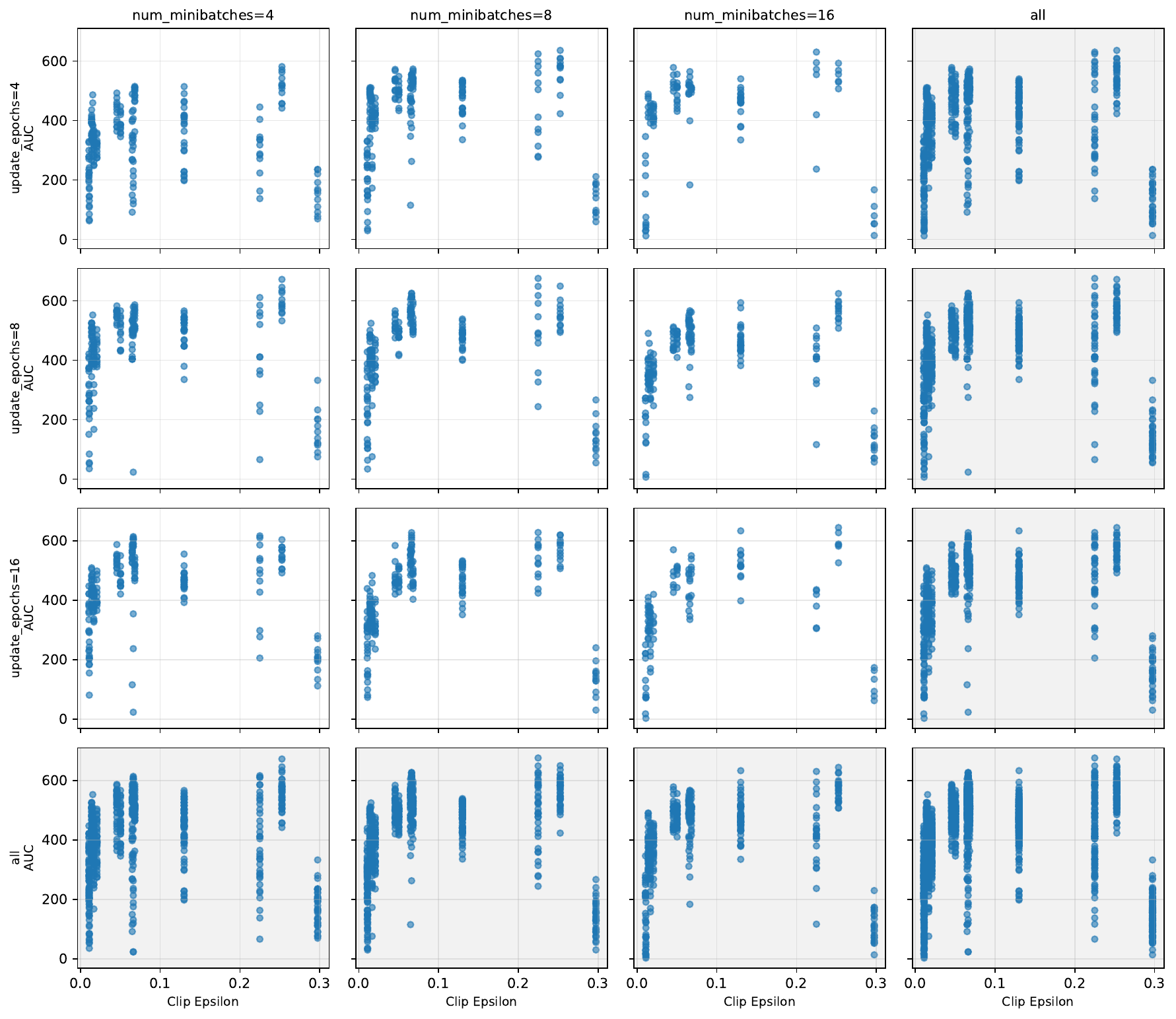}
    \caption{\textbf{IPPO (recurrent), Scratch Itch --- PPO clip-range sweep.} AUC return against the PPO clipping parameter $\epsilon$, where each marker is a single configuration. Panels are faceted by minibatch count (columns) and update epochs (rows), with \texttt{all} pooling over that factor.}
    \label{fig:sweep_ippo_rnn_si_clip}
\end{figure}

\begin{figure}[H]
    \centering
    \includegraphics[width=\linewidth]{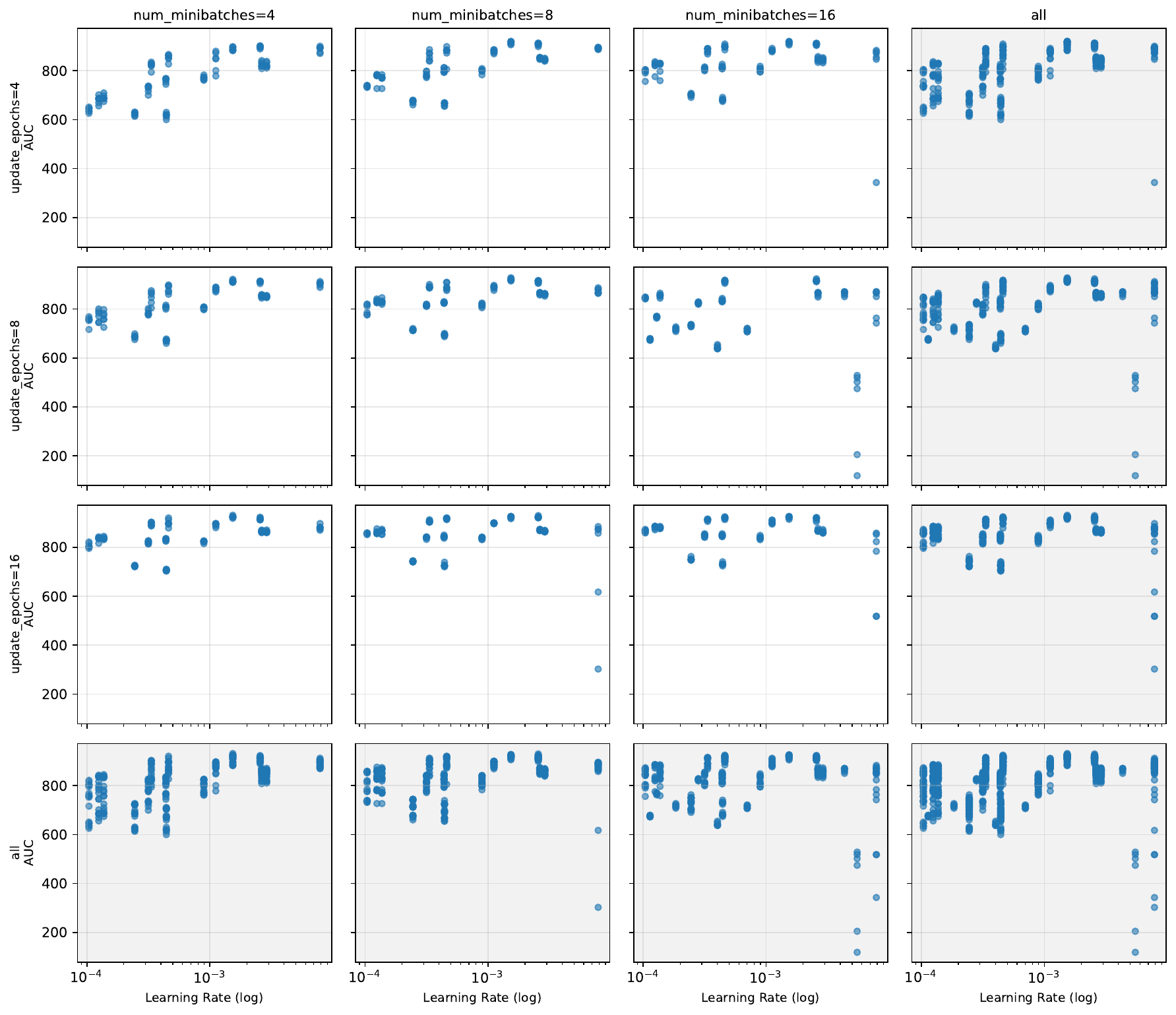}
    \caption{\textbf{MAPPO (feed-forward), Scratch Itch --- learning-rate sweep.} Each marker gives one configuration's AUC return as a function of the learning rate (log $x$-axis); higher is better. Panels are faceted by minibatch count (columns) and update epochs (rows), with \texttt{all} pooling over that factor.}
    \label{fig:sweep_mappo_ff_si_lr}
\end{figure}

\begin{figure}[H]
    \centering
    \includegraphics[width=\linewidth]{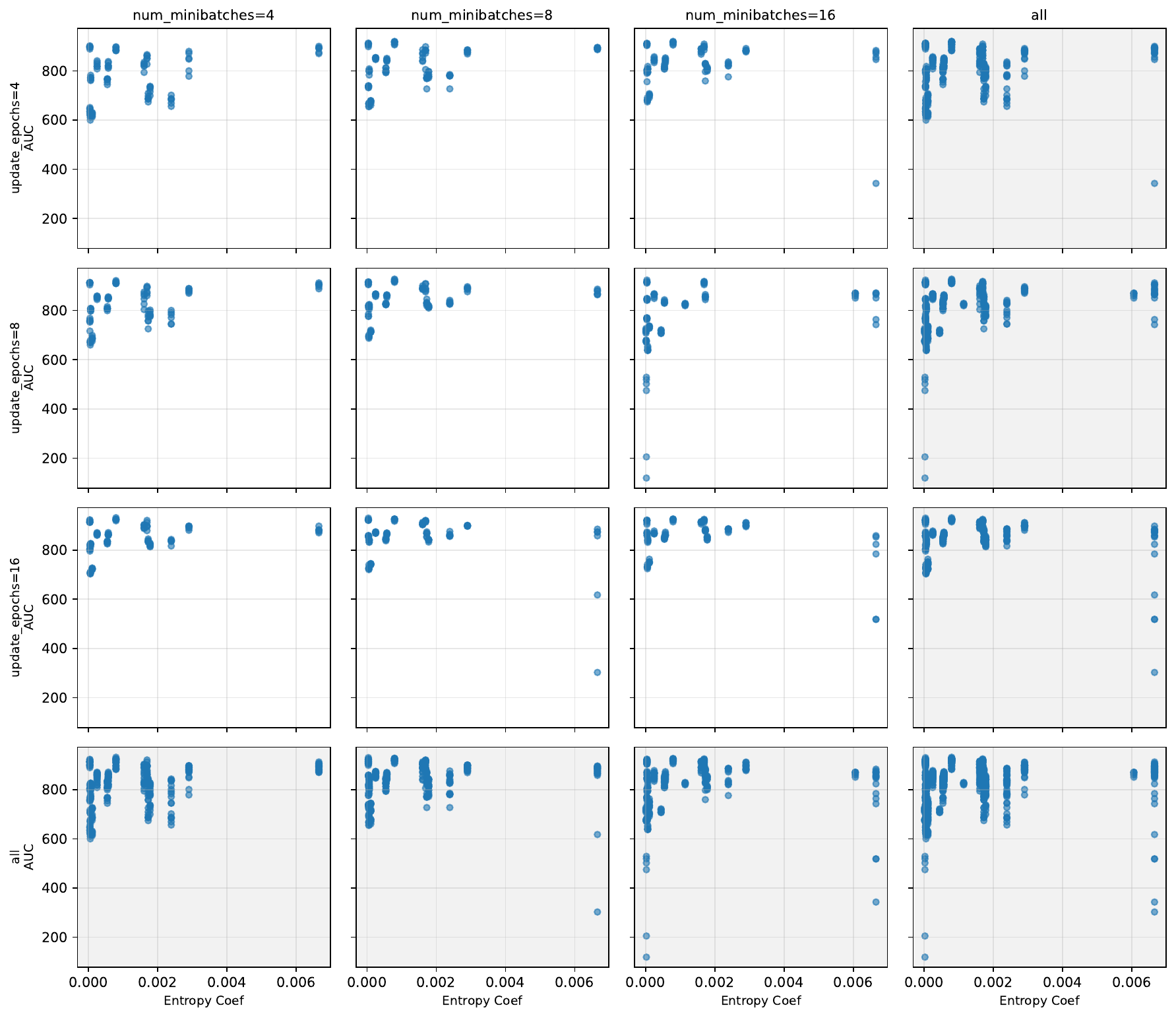}
    \caption{\textbf{MAPPO (feed-forward), Scratch Itch --- entropy-coefficient sweep.} AUC return against the entropy-regularisation coefficient, where each marker is a single configuration. Panels are faceted by minibatch count (columns) and update epochs (rows), with \texttt{all} pooling over that factor.}
    \label{fig:sweep_mappo_ff_si_ent}
\end{figure}

\begin{figure}[H]
    \centering
    \includegraphics[width=\linewidth]{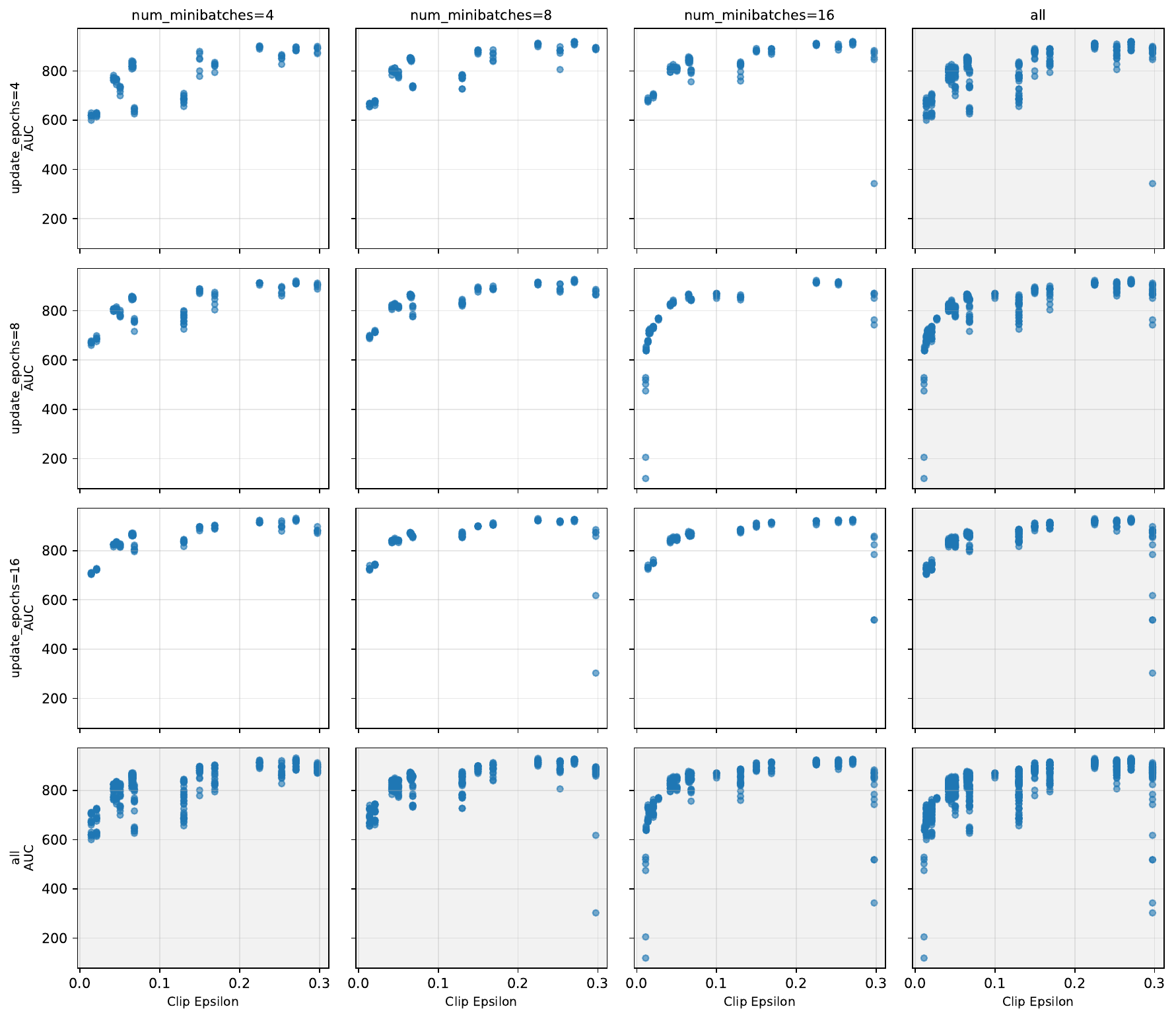}
    \caption{\textbf{MAPPO (feed-forward), Scratch Itch --- PPO clip-range sweep.} AUC return against the PPO clipping parameter $\epsilon$, where each marker is a single configuration. Panels are faceted by minibatch count (columns) and update epochs (rows), with \texttt{all} pooling over that factor.}
    \label{fig:sweep_mappo_ff_si_clip}
\end{figure}

\begin{figure}[H]
    \centering
    \includegraphics[width=\linewidth]{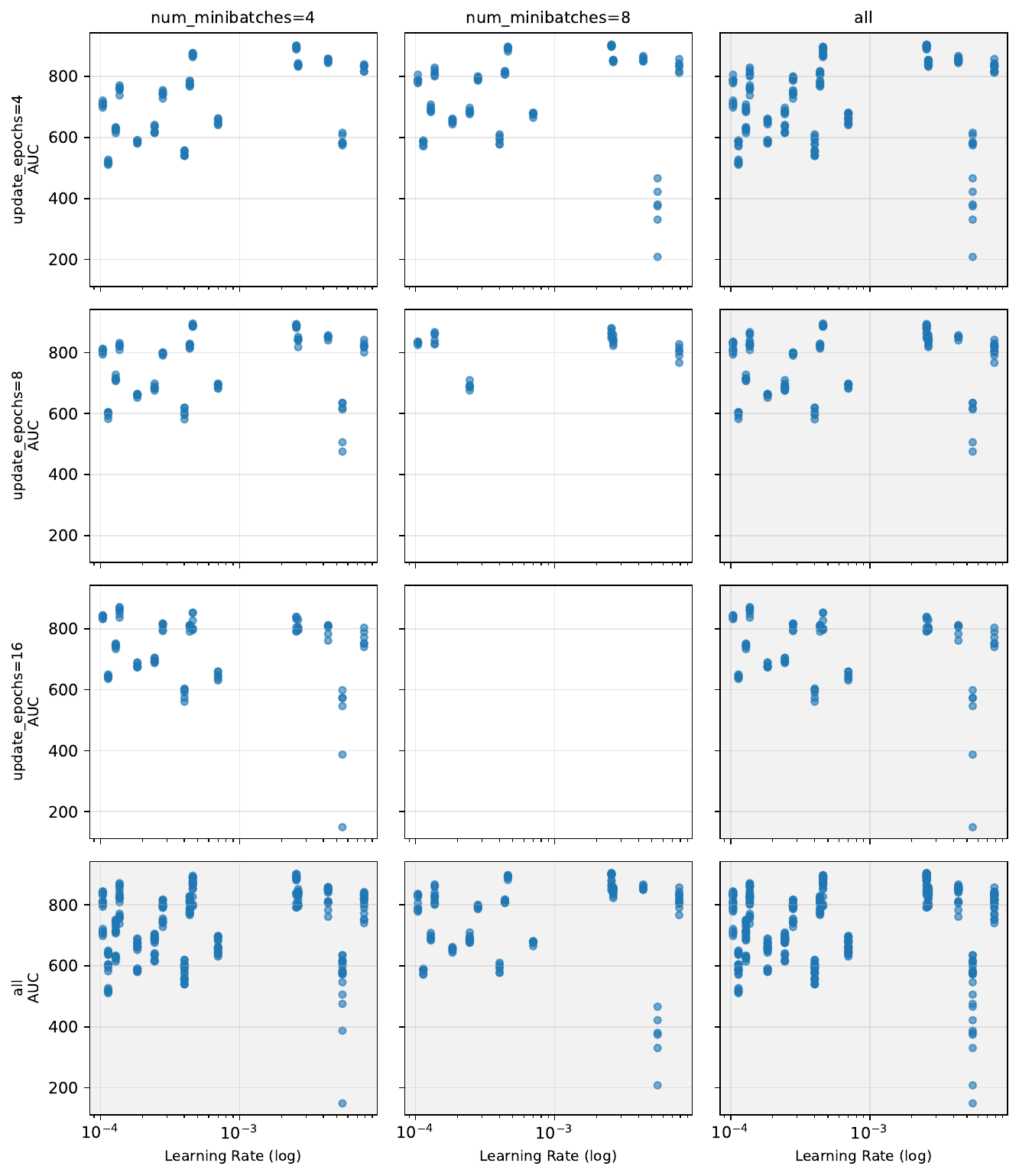}
    \caption{\textbf{MAPPO (recurrent), Scratch Itch --- learning-rate sweep.} Each marker gives one configuration's AUC return as a function of the learning rate (log $x$-axis); higher is better. Panels are faceted by minibatch count (columns) and update epochs (rows), with \texttt{all} pooling over that factor.}
    \label{fig:sweep_mappo_rnn_si_lr}
\end{figure}

\begin{figure}[H]
    \centering
    \includegraphics[width=\linewidth]{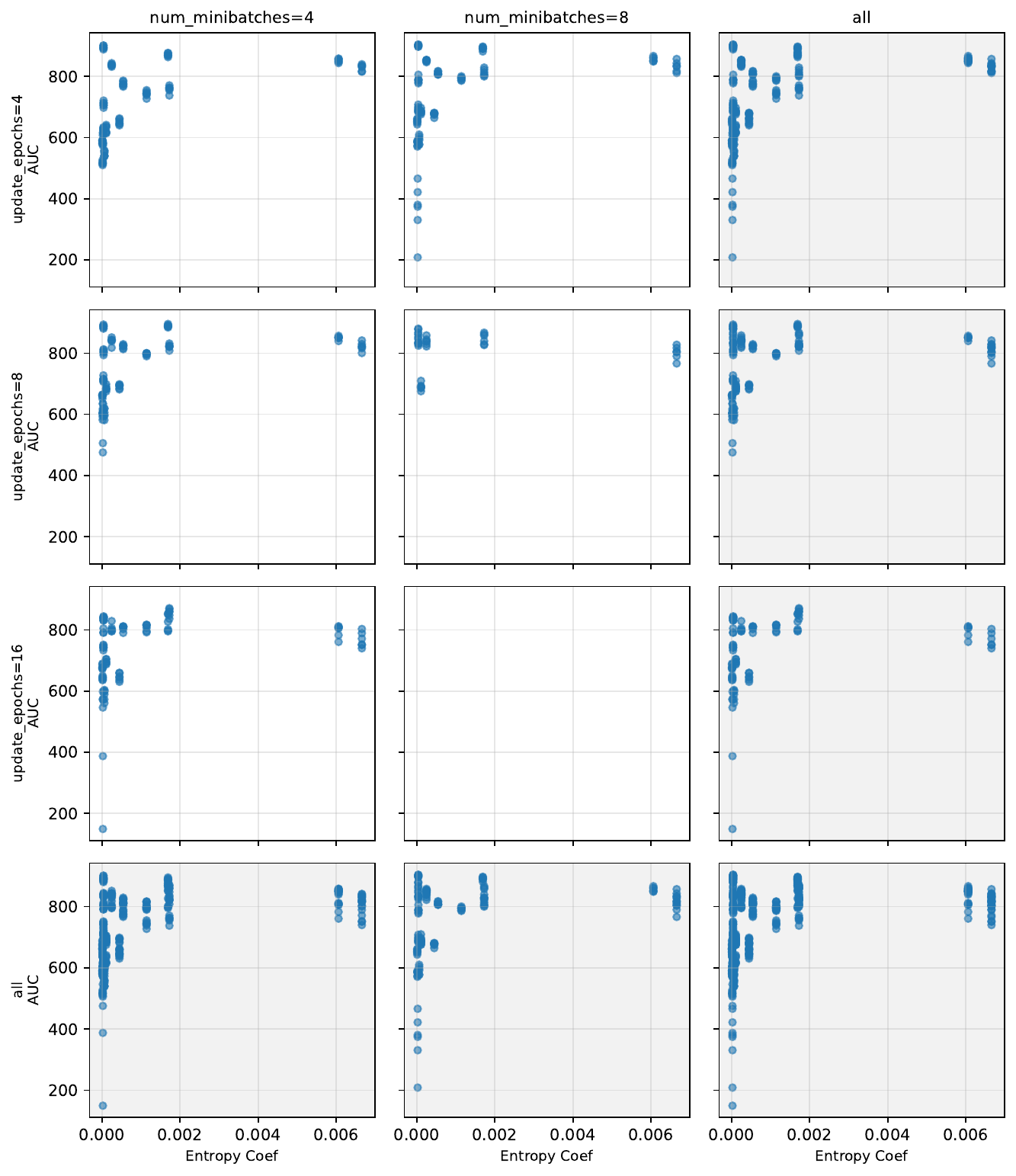}
    \caption{\textbf{MAPPO (recurrent), Scratch Itch --- entropy-coefficient sweep.} AUC return against the entropy-regularisation coefficient, where each marker is a single configuration. Panels are faceted by minibatch count (columns) and update epochs (rows), with \texttt{all} pooling over that factor.}
    \label{fig:sweep_mappo_rnn_si_ent}
\end{figure}

\begin{figure}[H]
    \centering
    \includegraphics[width=\linewidth]{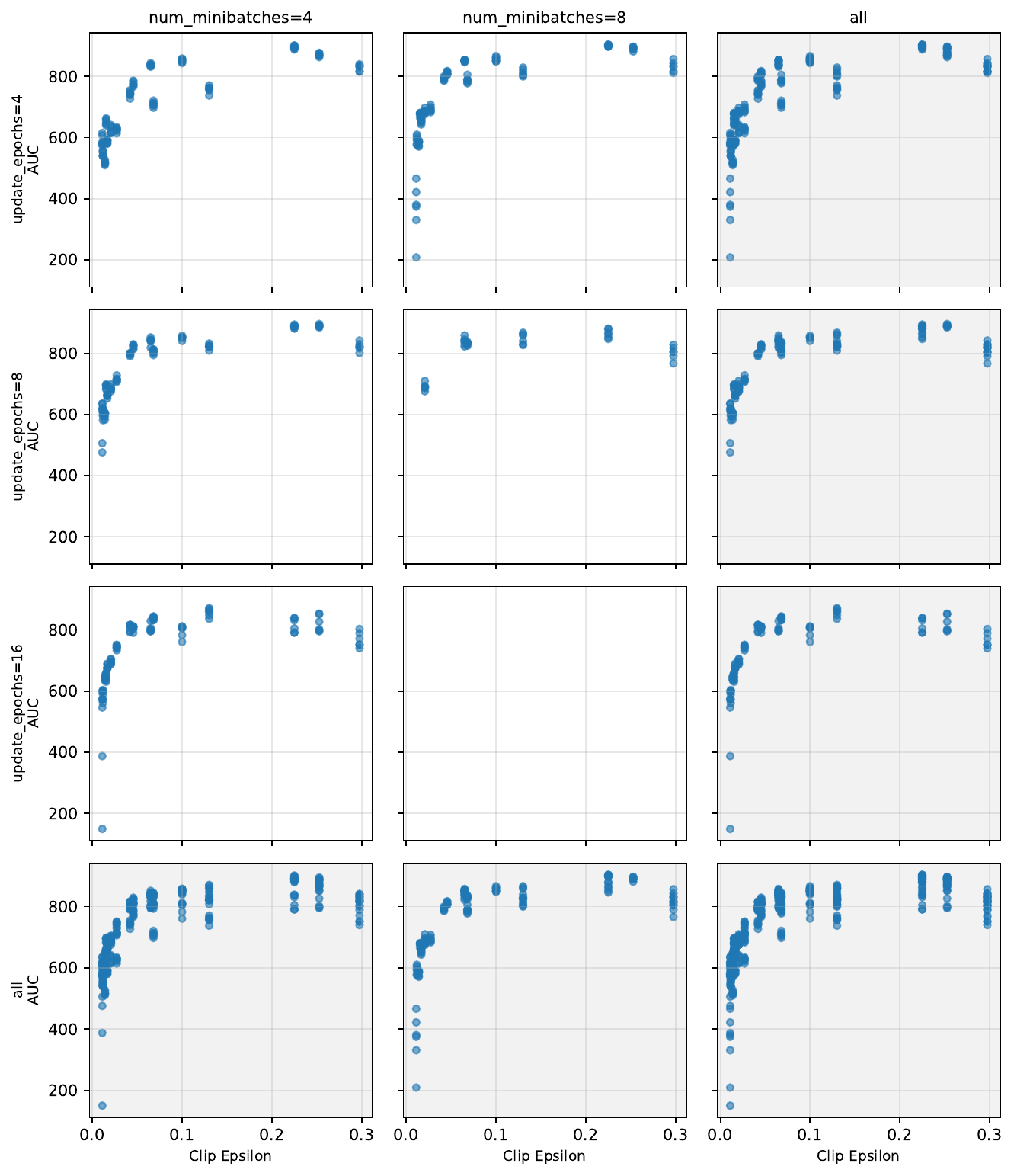}
    \caption{\textbf{MAPPO (recurrent), Scratch Itch --- PPO clip-range sweep.} AUC return against the PPO clipping parameter $\epsilon$, where each marker is a single configuration. Panels are faceted by minibatch count (columns) and update epochs (rows), with \texttt{all} pooling over that factor.}
    \label{fig:sweep_mappo_rnn_si_clip}
\end{figure}

\begin{figure}[H]
    \centering
    \includegraphics[width=\linewidth]{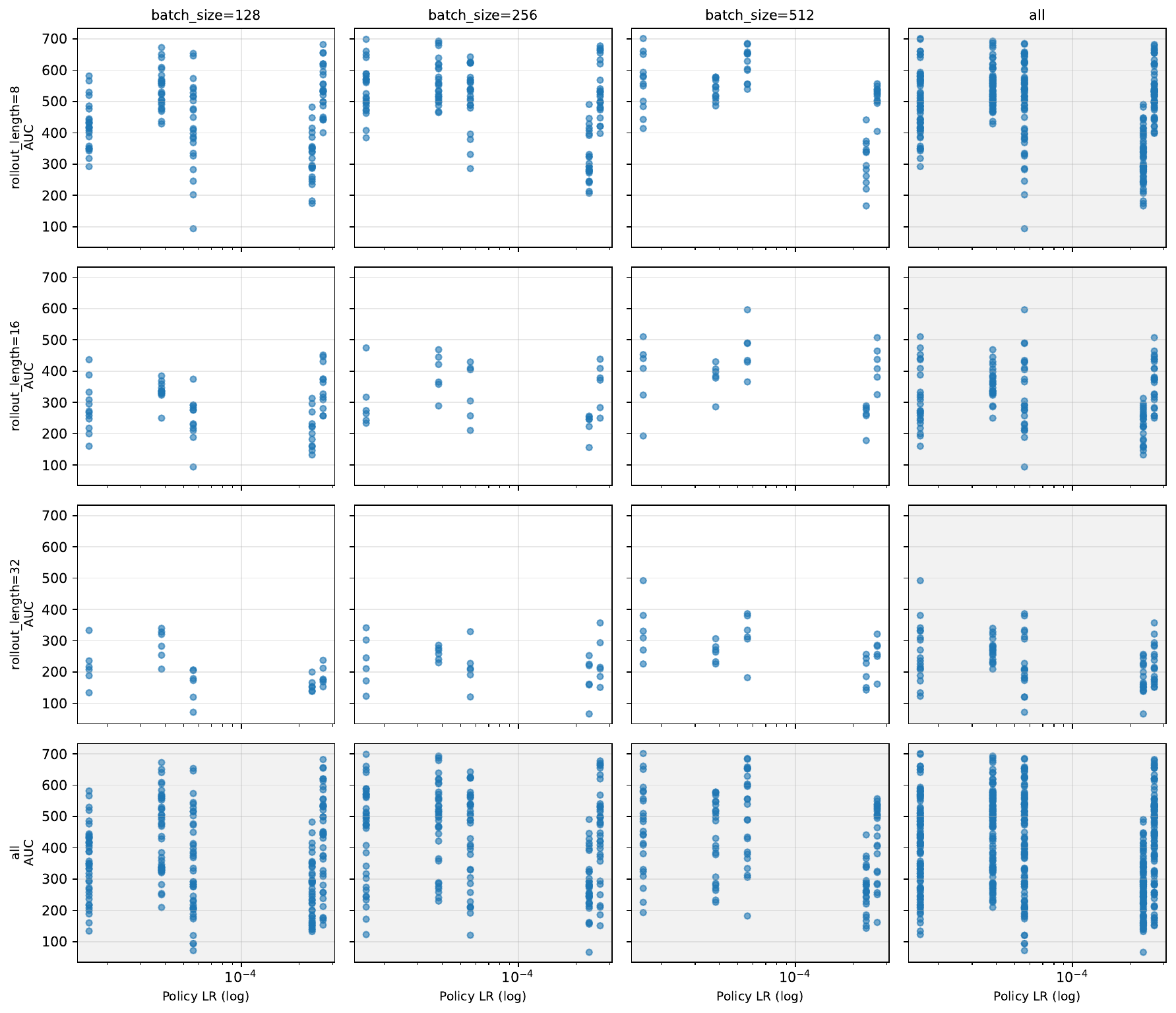}
    \caption{\textbf{MASAC (feed-forward), Scratch Itch --- policy (actor) learning-rate sweep.} Each marker gives one configuration's AUC return as a function of the actor learning rate (log $x$-axis); higher is better. Panels are faceted by batch size (columns) and rollout length (rows), with \texttt{all} pooling over that factor.}
    \label{fig:sweep_masac_ff_si_plr}
\end{figure}

\begin{figure}[H]
    \centering
    \includegraphics[width=\linewidth]{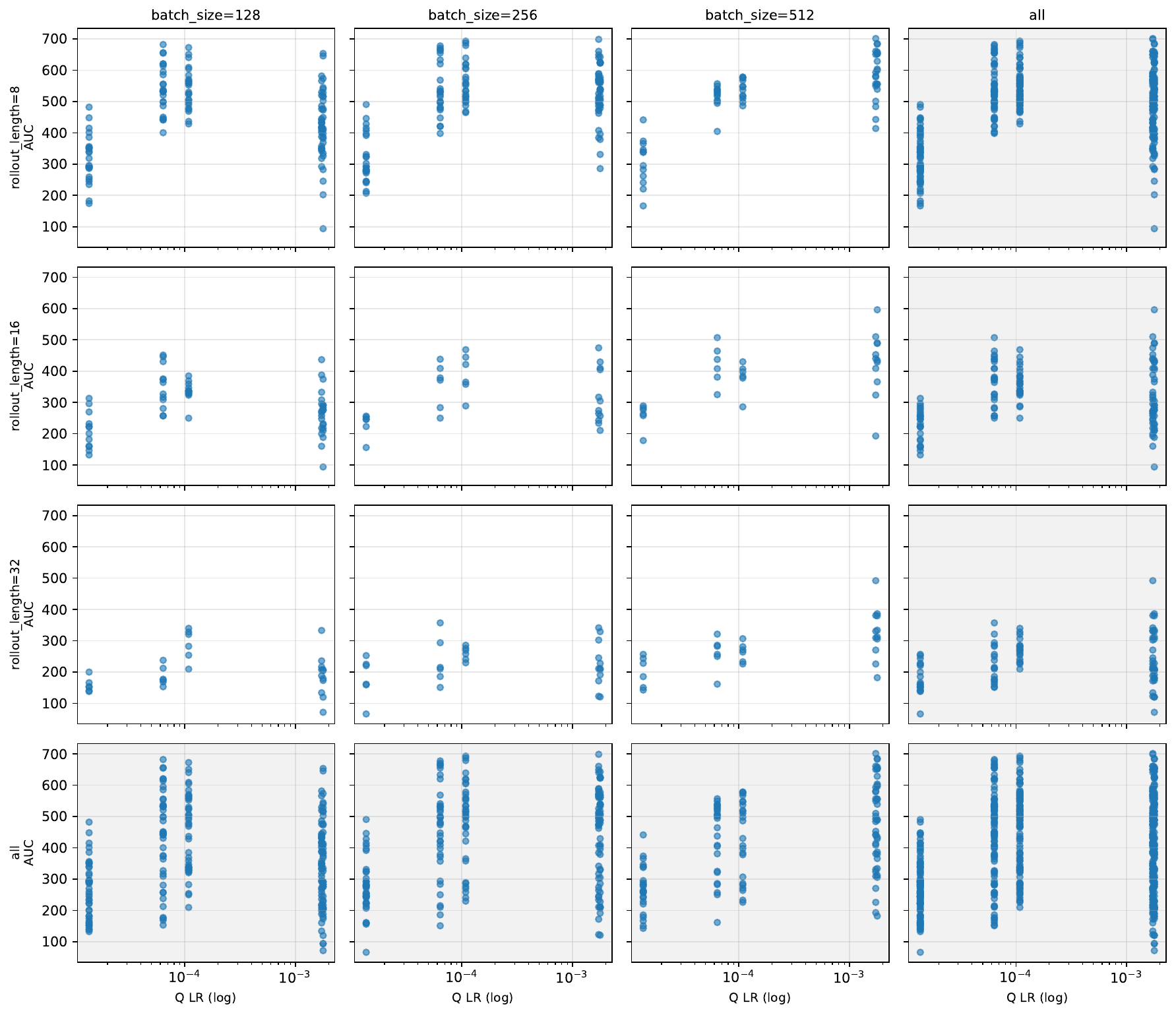}
    \caption{\textbf{MASAC (feed-forward), Scratch Itch --- critic ($Q$) learning-rate sweep.} AUC return against the critic learning rate (log $x$-axis), where each marker is a single configuration. Panels are faceted by batch size (columns) and rollout length (rows), with \texttt{all} pooling over that factor.}
    \label{fig:sweep_masac_ff_si_qlr}
\end{figure}

\begin{figure}[H]
    \centering
    \includegraphics[width=\linewidth]{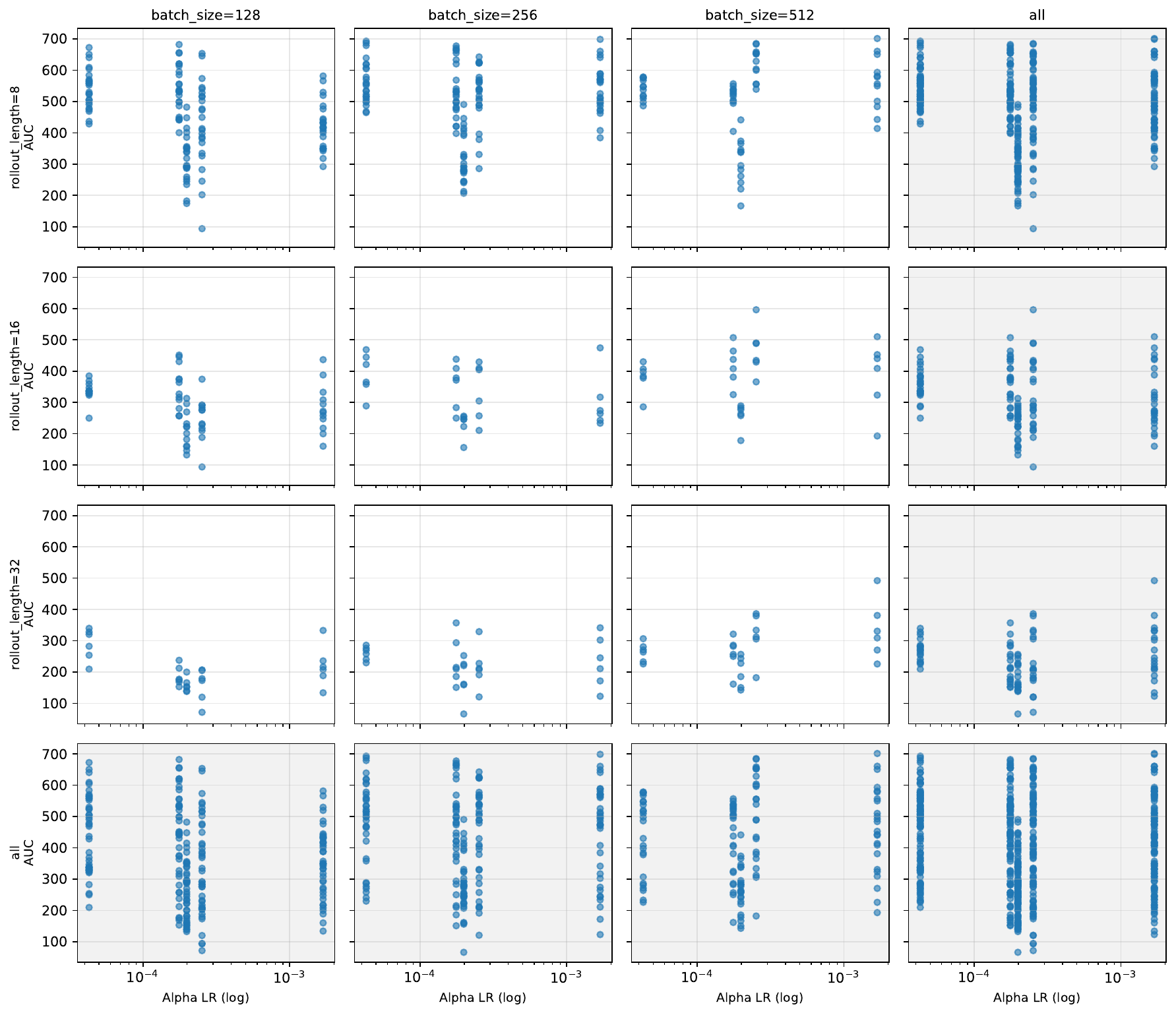}
    \caption{\textbf{MASAC (feed-forward), Scratch Itch --- temperature ($\alpha$) learning-rate sweep.} AUC return against the learning rate of the entropy-temperature $\alpha$ (log $x$-axis), where each marker is a single configuration. Panels are faceted by batch size (columns) and rollout length (rows), with \texttt{all} pooling over that factor.}
    \label{fig:sweep_masac_ff_si_alr}
\end{figure}

\begin{figure}[H]
    \centering
    \includegraphics[width=\linewidth]{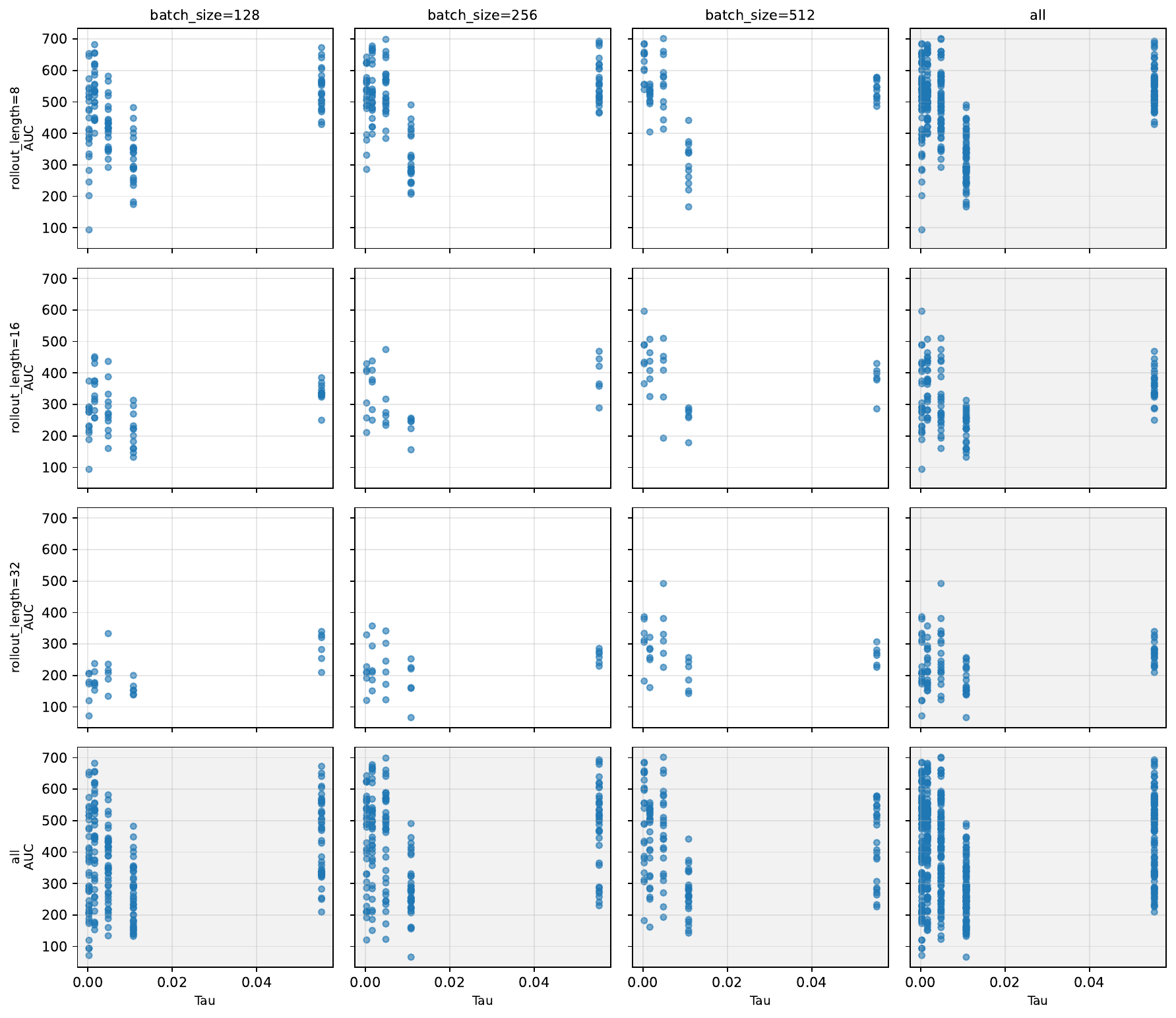}
    \caption{\textbf{MASAC (feed-forward), Scratch Itch --- target-smoothing ($\tau$) sweep.} AUC return against the target-network soft-update coefficient $\tau$, where each marker is a single configuration. Panels are faceted by batch size (columns) and rollout length (rows), with \texttt{all} pooling over that factor.}
    \label{fig:sweep_masac_ff_si_tau}
\end{figure}



\end{document}